\documentclass[10pt,twocolumn,letterpaper]{article}

\usepackage[pagenumbers]{cvpr} % To force page numbers, e.g. for an arXiv version

\definecolor{cvprblue}{rgb}{0.21,0.49,0.74}
\usepackage[pagebackref,breaklinks,colorlinks,allcolors=cvprblue]{hyperref}
\usepackage{amsmath,amssymb,amsfonts,booktabs,multirow,array}
\usepackage{graphicx,tikz,microtype,enumitem,algorithm,algorithmic}
\usepackage[table]{xcolor}
\usepackage{tabularx}
\usepackage{booktabs}
\usepackage{array}
\usepackage{pifont}
\newcommand{\sev}[1]{\textcolor{orange!80!black}{#1}}
\newcommand{\reg}[1]{\textcolor{blue!75!black}{#1}}
\newcommand{\inter}[1]{\textcolor{violet!80!black}{#1}}
\newcommand{\rulecolor}[1]{\textcolor{green!45!black}{#1}}

\def\confName{CVPR}
\def\confYear{2027}

\title{RecurIR: Learning When and How to Revise Restoration Policies via Recursive Self-Improvement}

\author{Hu Gao\\
Shanghai Jiao Tong University\\
{\tt\small gao\_h@sjtu.edu.cn}
\and
Haoran Zhang\\
Shanghai Jiao Tong University\\
{\tt\small zhr20040823@sjtu.edu.cn}
\and
Yulong Chen\\
Harbin Institute of Technology\\
{\tt\small llong\_c@hit.edu.cn}
\and
Lizhuang Ma\\
Shanghai Jiao Tong University\\
{\tt\small lzma@sjtu.edu.cn}
}

\begin{document}
\maketitle
\begin{abstract}
Agent-based image restoration has achieved promising performance by coordinating restoration tools to handle heterogeneous and coupled degradations. However, existing agents often improve their decisions through accumulated successful and failed executions, without explicitly identifying the conditions under which learned restoration strategies remain effective. Consequently, changes in degradation severity, spatial coverage, or interactions may invalidate previous decisions, while providing limited guidance on how these decisions should be revised. In this paper, we propose \textbf{RecurIR}, a recursive self-improvement framework that enables restoration agents to learn when and how to revise their decision rules. Specifically, we introduce a policy-hypothesis discovery mechanism to identify testable, condition-dependent rules through matched tool executions. We then design a rule-falsification-driven task evolution mechanism to generate targeted degradation variations, examine the validity of learned rules, and identify their applicability boundaries. Based on the resulting evidence, we develop a change-conditioned policy revision mechanism that learns how variations in degradation conditions should modify tool selection, execution order, operation strength, and processing regions. We further incorporate independent verification to retain effective revisions and use the resulting feedback to guide subsequent hypothesis discovery and task evolution. Through this recursive process, RecurIR progressively refines its restoration policies as degradation conditions change.  Extensive experiments demonstrate the effectiveness of the proposed framework.

\end{abstract}    
\section{Introduction}
Image restoration aims to recover high-quality visual content from degraded observations. Recent all-in-one image restoration (AiO-IR) methods have achieved promising performance by learning unified representations for different degradation types~\cite{CAPTNet10526271,Allrestor11367271}. Agentic image restoration has further demonstrated the benefits of coordinating specialized restoration tools through adaptive planning and execution~\cite{agenticir,hybridagentli2026hybrid,gao2026causal}. By adjusting restoration actions based on image characteristics and intermediate results, these agents can accommodate diverse degradation conditions. However, the effectiveness of a restoration strategy depends not only on degradation types but also on their severity, spatial distribution, and interactions. Consequently, a strategy that performs well under one condition may become less effective when these factors change, potentially requiring adjustments to tool selection, execution order, operation strength, or processing regions.

Existing restoration agents have explored various approaches to improving restoration decisions. Perception-guided planning and adaptive execution support flexible tool coordination~\cite{agenticir,hybridagentli2026hybrid}, while reinforcement learning enables the optimization of tool-use policies through restoration feedback~\cite{zhang2026tir,lu2026restorer1}. Experience-based approaches further improve decision-making by leveraging previous executions, refining restoration strategies, and reusing acquired knowledge~\cite{cui2026sear,zhuang2026evoir,gao2026causal,shao2026skillir}. More recent studies have also investigated predicting restoration outcomes before tool execution to reduce ineffective operations~\cite{zhang2026preact}. Nevertheless, the above approaches primarily focus on selecting effective restoration actions or learning from observed outcomes, while paying limited attention to actively identifying when learned decision rules become invalid and how they should be revised as degradation conditions change.

For example, an agent may learn from previous executions that denoising before deblurring yields favorable results. However, such experience does not necessarily indicate how the preferred strategy should change with noise severity, spatially varying blur, or localized degradation. Although experience-based memory supports tool selection and replanning, it may not explicitly capture the relationships between degradation conditions and restoration rules. Therefore, an agent should learn not only which strategies have been effective, but also under what conditions each rule should be selected, retained, or revised. By actively examining rule applicability and learning from verified revisions, the agent can progressively refine these condition-dependent relationships as degradation conditions change.

To address these challenges, we propose RecurIR, a recursive self-improvement framework that enables restoration agents to autonomously discover, challenge, and revise their decision rules. Specifically, we introduce a policy-hypothesis discovery mechanism to identify testable, condition-dependent relationships between degradation characteristics and restoration decisions through controlled tool comparisons. We then design a rule-falsification-driven task evolution mechanism to generate targeted variations in degradation severity, spatial distribution, and interactions, allowing the agent to challenge learned rules and identify their applicability boundaries. Based on the resulting evidence, we develop a change-conditioned policy revision mechanism to learn how restoration strategies should be adjusted in terms of tool selection, execution order, operation strength, and processing regions. To ensure reliable self-improvement, we further introduce independent transfer and retention tests to verify policy revisions and reject ineffective updates. The verified revisions subsequently guide hypothesis discovery and task evolution in the following rounds, forming a recursive process that progressively improves both restoration decisions and the strategies used to revise them.

The main contributions of this paper are summarized as follows:

\begin{itemize}
    \item We propose RecurIR, a recursive self-improvement framework that enables restoration agents to autonomously discover, challenge, and revise decision rules as degradation conditions change. Extensive experiments demonstrate its effectiveness across diverse restoration scenarios.

    \item We introduce a policy-hypothesis discovery mechanism that identifies testable relationships between degradation conditions and restoration decisions.

    \item We develop a rule-falsification-driven task evolution mechanism that generates targeted degradation variations to challenge learned decision rules and identify their limitations.

    \item We design a change-conditioned policy revision mechanism that adapts restoration decisions to changing degradation conditions and progressively improves the revision process through verified feedback.
\end{itemize}

\section{Related Work}

\subsection{Model-based Image Restoration}

Image restoration has evolved from prior-driven formulations to data-driven representation learning. Traditional approaches rely on explicit structural assumptions and degradation models to constrain image recovery~\cite{10558778,11342300}, while deep learning methods acquire restoration priors from data through increasingly expressive architectures~\cite{gao2025mixed,gao2026emphasizing,LSSRgao2024learning}. Task-specific methods have improved restoration quality through local--global feature interaction~\cite{ALGgao2024learning}, spatial dependency modeling~\cite{liu2024xyscannet}, and spatial--frequency representations~\cite{FSNet,FDTANetgao2025frequency}. Other studies employ predictive filtering~\cite{efderainguo2025efficientderain+}, explicit image priors~\cite{PGH2Netisu2025prior}, and diffusion-based generation~\cite{upid10.1145/3664647.3680560,diffunmix10656884} to enhance structural and detail recovery. Despite their effectiveness, these methods are generally designed for specific restoration tasks, limiting their adaptability to diverse degradation conditions.

All-in-one image restoration extends these advances by addressing multiple degradations within a unified framework. AdaIR~\cite{cui2025adair} explores adaptive spatial--frequency representations, while Perceive-IR~\cite{Perceive-IR10990319} incorporates degradation categories and severity to guide restoration. BaryIR~\cite{baryir11417902} aligns heterogeneous degradation distributions to enhance cross-task adaptation, and AllRestorer~\cite{Allrestor11367271} integrates visual and textual information to characterize complex degradation conditions. Recent approaches further leverage vision-language guidance to improve unified restoration~\cite{VLUNetZeng_2025_CVPR,autodir10.1007/978-3-031-73661-2_19,DefusionLuo_2025_CVPR}.

Despite these advances, model-based methods rely largely on learned representations and input-dependent conditioning, limiting their flexibility under complex and interacting degradations. Agent-based approaches offer greater adaptability through dynamic tool coordination and feedback-driven decision-making.

\subsection{Agent-based Image Restoration}

The development of large language models~\cite{LLMbrown2020language,LLMroziere2023code,LLMtouvron2023llama} and multimodal reasoning models~\cite{MLLMli2023blip,MLLMliu2023visual,MLLMsingh2025openai} has enabled image restoration systems to perform sequential decision-making through adaptive tool coordination. These systems analyze degraded observations, select specialized restoration tools, and adjust subsequent operations based on intermediate results. RestoreAgent~\cite{RestoreAgent} explores vision-language-guided restoration planning, while AgenticIR~\cite{agenticir} combines degradation perception, tool scheduling, and feedback-driven replanning. MAIR~\cite{MAIRjiang2026multi} coordinates restoration decisions across multiple agents, and HybridAgent~\cite{hybridagentli2026hybrid} integrates different planning strategies to balance restoration quality and computational efficiency. Recent approaches further employ policy learning to optimize tool selection and reduce unnecessary executions~\cite{zhang2026tir,lu2026restorer1}.
Another research direction focuses on improving restoration decisions through experience reuse and adaptive knowledge refinement. SEAR~\cite{cui2026sear} combines deliberate search with evolving episodic memory to guide subsequent tool execution, while EvoIR-Agent~\cite{zhuang2026evoir} organizes previous restoration experiences to support future decisions. Causal-AgentIR~\cite{gao2026causal} further introduces structured causal memory to facilitate adaptive restoration reasoning. SkillIR~\cite{shao2026skillir} develops context-dependent restoration skills and refines their applicability using execution feedback. Meanwhile, PreActIR~\cite{zhang2026preact} explores tool-effect prediction to estimate restoration outcomes before execution.

These advances have improved the flexibility and effectiveness of restoration agents through adaptive planning, policy optimization, and experience reuse. However, existing approaches provide relatively limited explicit modeling of how changes in degradation conditions should translate into modifications of restoration decisions. Although some methods refine the applicability of learned actions or skills, systematically challenging conditional decision rules and learning corresponding policy revisions remain less explored. RecurIR addresses this problem by learning how degradation variations affect restoration decisions and using the resulting evidence to revise existing policies.
\section{Method}
\label{sec:method}

\begin{figure}
    \centering
    \includegraphics[width=\linewidth]{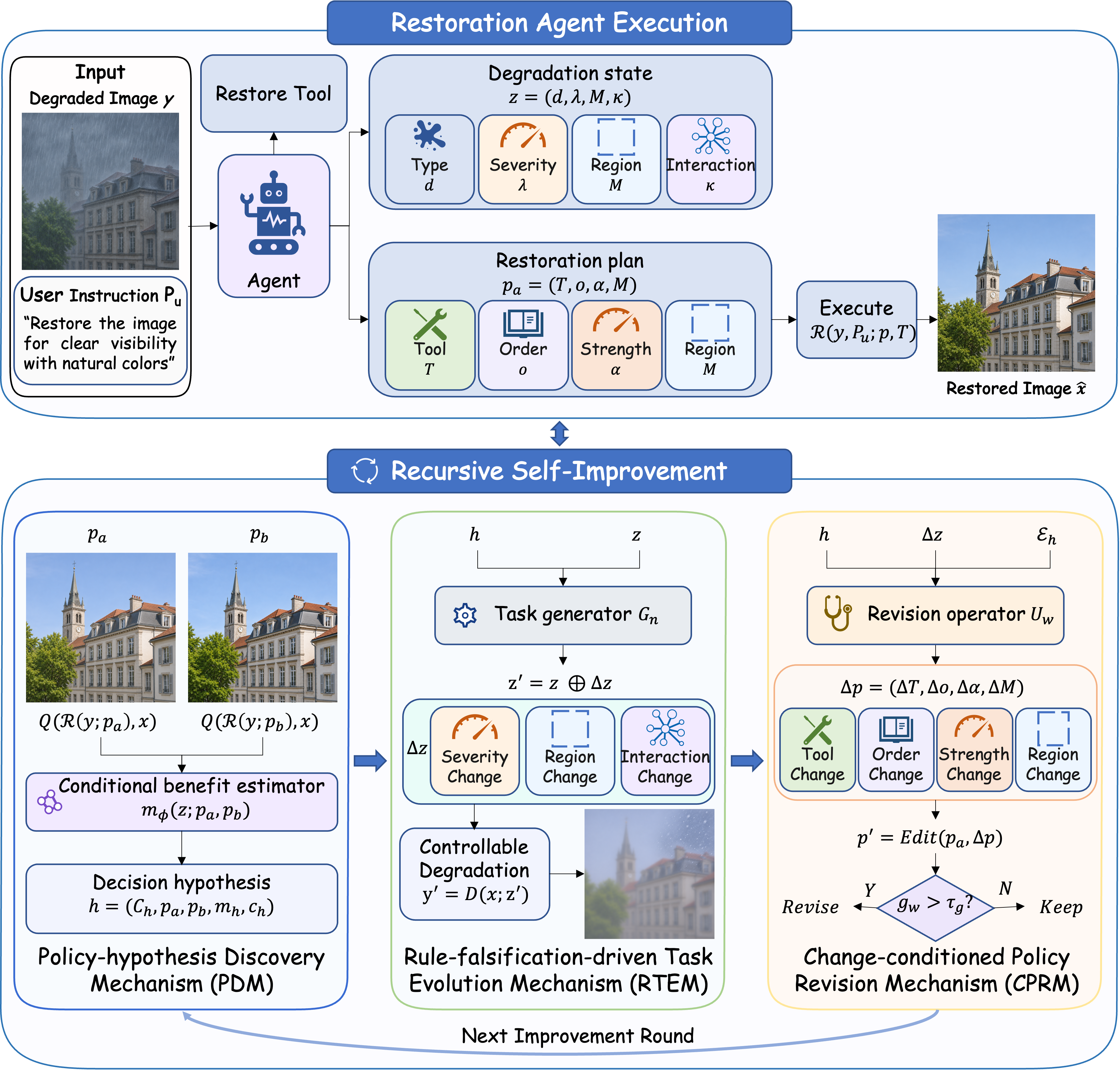}
    \caption{Overview of the proposed RecurIR framework. \textbf{Top:} A general agentic image restoration pipeline. \textbf{Bottom:} The proposed recursive self-improvement framework, consisting of policy-hypothesis discovery (PDM), rule-falsification-driven task evolution (RTEM), and change-conditioned policy revision (CPRM). }
    \label{fig:framework}
\end{figure}

\subsection{Overall Pipeline}
\label{sec:overview}

We propose RecurIR, a recursive self-improvement framework that enables restoration agents to autonomously discover, challenge, and revise their decision rules as degradation conditions change. As illustrated in Figure~\ref{fig:framework}, RecurIR consists of three interconnected components: policy-hypothesis discovery mechanism (PDM), rule-falsification-driven task evolution mechanism (RTEM), and change-conditioned policy revision mechanism (CPRM).

Given a degraded image $y$ and the user instruction $P_u$, an agent follows a policy $\pi_{\theta}$ to select a restoration plan $p$ from an available toolbox $\mathcal{T}=\{T_1,\ldots,T_N\}$. The plan specifies the selected tools, execution order, operation parameters, and processing regions. The restored image is obtained by:
\begin{equation}
    \hat{x}=\mathcal{R}(y, P_u;p,\mathcal{T}),
    \label{eq:restoration}
\end{equation}
where $\mathcal{R}$ denotes the execution of the restoration plan.
To characterize the conditions governing restoration decisions, we represent the degradation state as $ z=(d,\lambda,M,\kappa)$, where $d$, $\lambda$, $M$, and $\kappa$ describe degradation types, severity, spatial distribution, and interactions, respectively. These conditions are available during controlled degradation generation and estimated from image observations during inference.

RecurIR operates through multiple self-improvement rounds. At round $k$, the current agent discovers decision hypotheses $\mathcal{H}_k$ from matched restoration comparisons. A task generator then constructs degradation variations to challenge these hypotheses and identify conditions under which they become invalid. The resulting evidence guides a revision policy that learns how changes in degradation conditions should modify restoration decisions. Candidate updates are independently verified before being accepted. The revised agent subsequently discovers new hypotheses and generates new learning tasks, allowing both restoration decisions and their revision mechanisms to improve recursively.

\subsection{Policy-Hypothesis Discovery}
\label{sec:hypothesis}

Restoration experiences provide evidence about which operations perform well on previously encountered images, but do not necessarily reveal the conditions governing their effectiveness. We therefore introduce a policy-hypothesis discovery mechanism (PDM) that explicitly learns conditional relationships between degradation characteristics and restoration decisions.
Given a degraded observation $y$ and two candidate restoration plans $p_a$ and $p_b$, their relative restoration benefit is defined as:
\begin{equation}
\Delta Q(y;p_a,p_b)
=Q(\mathcal{R}(y;p_a),x)-Q(\mathcal{R}(y;p_b),x),
\label{eq:action_benefit}
\end{equation}
where $x$ denotes the reference image and $Q$ represents a restoration quality measure. Reference images are used to construct training supervision and are not required during inference. A positive $\Delta Q$ indicates that $p_a$ is preferable to $p_b$ under the current observation. However, this relationship may change with degradation conditions. To capture such dependencies, we introduce a conditional benefit estimator $m_{\phi}(z;p_a,p_b) =f_{\phi}(z,p_a,p_b)$, which predicts the relative effectiveness of competing restoration plans given degradation state $z$.
The estimator is optimized using matched-state restoration comparisons:
\begin{equation}
\mathcal{L}_{\mathrm{hyp}}
=\mathbb{E}_{\mathcal{E}} \Big[
\mathcal{L}_{\mathrm{BCE}}
\left(
\sigma(m_{\phi}/\tau),
\mathbb{I}[\Delta Q>0]
\right) +\lambda_m|m_{\phi}-\Delta Q|
\Big],
\label{eq:hypothesis_loss}
\end{equation}
where $\mathcal{E}$ contains matched comparison evidence, $\sigma(\cdot)$ denotes the sigmoid function, and $\tau$ is a temperature parameter. The classification term learns action preferences, while the regression term captures their relative benefits.
Based on the learned conditional relationships, a restoration decision hypothesis is represented as:
\begin{equation}
    h=(\mathcal{C}_h,p_a,p_b,m_h,c_h),
    \label{eq:policy_hypothesis}
\end{equation}
where $\mathcal{C}_h$ denotes the applicability region, $m_h$ represents the predicted benefit relationship, and $c_h$ indicates hypothesis confidence.
The applicability region defines the degradation conditions under which the preferred restoration decision is expected to remain effective, while hypothesis confidence reflects the consistency and reliability of matched comparisons. Each hypothesis captures a testable relationship between degradation conditions and restoration decisions, specifying both the preferred strategy and its applicability. This representation enables the agent to assess the validity of previously learned rules under current degradation conditions and guides subsequent rule selection and revision.

\subsection{Rule-Falsification-Driven Task Evolution}
\label{sec:task_evolution}

A learned decision hypothesis may become inappropriate when degradation conditions change. To proactively identify such failures, we introduce a rule-falsification-driven task evolution mechanism (RTEM) that generates targeted degradation variations to examine the validity of existing hypotheses.
Given a hypothesis $h$ associated with degradation state $z$, the task generator produces a modified condition:
\begin{equation}
    z'=z\oplus\Delta z,\quad
    \Delta z\sim G_{\eta_k}(\cdot\mid z,h),
    \label{eq:task_generation}
\end{equation}
where $\Delta z$ represents controlled changes in degradation severity, spatial distribution, or interactions, and $G_{\eta_k}$ denotes the task generation policy at improvement round $k$.
The corresponding observation is generated as:
\begin{equation}
    y'=\mathcal{D}(x;z'),
    \label{eq:generated_observation}
\end{equation}
where $\mathcal{D}$ denotes a controllable degradation process. Using the same underlying image allows restoration decisions to be evaluated under matched content conditions. While conventional hard-example mining typically aims to maximize reconstruction error, RecurIR focuses on identifying degradation variations that reveal inconsistencies between learned decision hypotheses and observed restoration outcomes.

Consider a hypothesis predicting that $p_a$ is preferable to $p_b$. A generated task provides evidence against this hypothesis when the predicted preference is confidently supported but contradicted by the actual restoration results.
We define the falsification reward as:
\begin{equation}
r_{\mathrm{fal}}(h,z')
=\mathbb{I}[z'\in\mathcal{C}_h] \cdot \mathbb{I}[m_h(z')>\tau_m]  \cdot [-\Delta Q(y';p_a,p_b)-\epsilon]_+,
\label{eq:falsification_reward}
\end{equation}
where $[u]_+=\max(0,u)$, $\tau_m$ denotes a prediction margin, and $\epsilon$ specifies the minimum meaningful improvement.
This reward favors conditions under which the hypothesis confidently prefers $p_a$, although $p_b$ achieves better restoration quality. By restricting falsification to the claimed applicability region, the mechanism distinguishes genuine rule contradictions from conditions outside the scope of the hypothesis.
To further identify where action preferences change, we introduce a boundary exploration score:
\begin{equation}
    b_h(z')=
    \exp\left(
    -\frac{|m_h(z')|}{\tau_b}
    \right),
    \label{eq:boundary_score}
\end{equation}
where $\tau_b$ controls the sensitivity to uncertain action preferences.
The task generator is optimized by combining rule falsification and boundary exploration:
\begin{equation}
\max_{\eta}
\mathbb{E}_{\Delta z\sim G_{\eta}}
\Big[
r_{\mathrm{fal}}(h,z')
+\lambda_b b_h(z') - \lambda_c\mathcal{C}(\Delta z)
\Big],
\label{eq:task_objective}
\end{equation}
where $\mathcal{C}$ penalizes implausible degradation variations and excessive evaluation costs.
The evaluated tasks provide evidence about where existing rules fail and how their applicability changes. This evidence is used to update both the decision hypotheses and the subsequent task generation policy. The generated counterexamples are further used to refine the applicability regions of existing hypotheses. Specifically, conditions that contradict a learned preference provide evidence for restricting its validity, while boundary samples help identify where alternative restoration strategies become preferable. The refined hypotheses and their associated evidence are subsequently used to supervise policy revision and guide the next round of task evolution.

\subsection{Change-Conditioned Policy Revision}
\label{sec:policy_revision}

Identifying an invalid decision hypothesis does not directly determine how the restoration policy should be corrected. We therefore introduce a change-conditioned policy revision mechanism (CPRM) that explicitly learns how variations in degradation conditions should modify restoration decisions.
Given a hypothesis $h$ associated with reference condition $z$ and a modified condition $z'$, their difference is represented as $\Delta z=\operatorname{Diff}(z',z)$.
To adapt existing restoration strategies to changing degradation conditions, we introduce a revision operator that predicts the required adjustments:
\begin{equation}
\begin{aligned}
    \Delta p
    &=U_{\omega_k}(h,\Delta z,\mathcal{E}_h),\\
    p'
    &=\operatorname{Edit}(p_a,\Delta p),
\end{aligned}
\label{eq:revision_operator}
\end{equation}
where $\mathcal{E}_h$ contains previously verified evidence associated with the hypothesis. The predicted revision is represented as $ \Delta p = (\Delta T,\Delta o,\Delta\alpha,\Delta M)$,
where the four components correspond to adjustments in tool selection, execution order, operation strength, and processing regions, respectively.
This enables the agent to learn how an existing strategy should be modified when degradation conditions vary, allowing the agent to adapt existing decisions beyond experience retrieval. To construct revision supervision, candidate restoration plans are evaluated under the modified condition. The best plan within the evaluated candidate set is selected as:
\begin{equation}
    p^*=
    \arg\max_{p\in\mathcal{P}(y')}
    Q(\mathcal{R}(y';p),x),
    \label{eq:reference_plan}
\end{equation}
where $\mathcal{P}(y')$ contains feasible candidate plans. The target revision is obtained by comparing $p^*$ with the original plan $p_a$. Meanwhile, we introduce a revision gate $g_{\omega}$ to determine whether the original plan requires modification. Specifically, the gate employs a lightweight two-layer MLP followed by a sigmoid function. It takes the encoded decision hypothesis $h$, degradation change $\Delta z$, and aggregated evidence $\mathcal{E}_h$ as inputs, and outputs a revision probability $g_{\omega}\in(0,1)$. The revision target is defined as:

\begin{equation}
r^*=
\mathbb{I}\left[
Q(\mathcal{R}(y';p^*),x)
-
Q(\mathcal{R}(y';p_a),x)
>\epsilon
\right].
\label{eq:revision_target}
\end{equation}

The revision mechanism is optimized through:
\begin{equation}
\begin{aligned}
\mathcal{L}_{\mathrm{rev}}
=&
\mathbb{E}
\Big[
\mathcal{L}_{\mathrm{BCE}}(g_{\omega},r^*)
-\lambda_p r^*
\log P_{\omega}(\Delta p^*\mid h,\Delta z,\mathcal{E}_h)
\Big]
\\
&+\lambda_r\mathcal{L}_{\mathrm{ret}},
\label{eq:revision_loss} 
\end{aligned}
\end{equation}
where $\Delta p^*$ denotes the target plan adjustment and $\mathcal{L}_{\mathrm{ret}}$ is a retention regularizer that discourages unnecessary changes to previously reliable decisions. 
During inference, the revised restoration plan is determined by:
\begin{equation}
p_{\mathrm{new}}=
\begin{cases}
\operatorname{Edit}(p_a,\Delta p),
&g_{\omega}>\tau_g,\\
p_a,
&\mathrm{otherwise},
\end{cases}
\label{eq:adaptive_revision}
\end{equation}
where $\tau_g$ is the revision threshold. The current degradation state is estimated from the input image, while its reference condition is obtained from the selected hypothesis. 
To prevent unreliable revisions from affecting previously acquired capabilities, candidate policy updates are evaluated on independent transfer and retention validation sets.
Let $\pi_k$ and $\tilde{\pi}_{k+1}$ denote the current and candidate updated policies. Their performance changes are measured as:
\begin{equation}
\begin{aligned}
\Delta Q_{\mathrm{tr}}
&=\overline{Q}_{\mathrm{tr}}(\tilde{\pi}_{k+1})
-\overline{Q}_{\mathrm{tr}}(\pi_k),\\
\Delta Q_{\mathrm{ret}}
&=\overline{Q}_{\mathrm{ret}}(\tilde{\pi}_{k+1})
-\overline{Q}_{\mathrm{ret}}(\pi_k),
\end{aligned}
\label{eq:verification_metrics}
\end{equation}
where the transfer set evaluates decisions under changed degradation conditions, while the retention set measures performance on previously learned conditions.
The candidate update is accepted when
$
\mathbb{I}
\left[
\Delta Q_{\mathrm{tr}}\geq\delta_{\mathrm{tr}}
\land
\Delta Q_{\mathrm{ret}}\geq-\epsilon_{\mathrm{ret}}
\right]$, $\delta_{\mathrm{tr}}$ denotes the required improvement and $\epsilon_{\mathrm{ret}}$ specifies the allowable performance reduction.
Updates failing these criteria are rejected. Successful revisions update the agent policy and its revision mechanism, which subsequently guide new hypothesis discovery and task generation. Through this process, RecurIR progressively improves not only restoration decisions but also its ability to determine when and how those decisions should be revised.

\section{Experiments}
\label{sec:exp}
In this section, we present qualitative and quantitative comparisons, followed by ablation studies to evaluate the effectiveness of the proposed method. Due to space limitations, detailed experimental settings and additional results are provided in the supplementary material~\ref{sec:exset} and~\ref{sec:moreex}.

\begin{table*}
\centering
\caption{Quantitative comparison of all-in-one restoration models and agent-based methods across five restoration tasks. Denoising is evaluated at a noise level of $\sigma=15$. The best and second-best results are highlighted in bold and underlined, respectively.}
\label{tb:allrshbn}
    \resizebox{\linewidth}{!}{
\begin{tabular}{c|c|cccccccccc|cc}
    \hline
   \multirow{2}{*}{Type} & \multirow{2}{*}{Methods} & \multicolumn{2}{c}{Deraining} & \multicolumn{2}{c}{Desnowing} & \multicolumn{2}{c}{Dehazing}& \multicolumn{2}{c}{Deblurring}& \multicolumn{2}{c|}{Denoising}  & \multicolumn{2}{c}{Average} 
    \\
   & &PSNR $\uparrow$ &  SSIM $\uparrow$  &PSNR $\uparrow$ &SSIM $\uparrow$ & PSNR $\uparrow$&SSIM $\uparrow$ &PSNR $\uparrow$ & SSIM $\uparrow$&PSNR $\uparrow$ & SSIM $\uparrow$&PSNR $\uparrow$ & SSIM $\uparrow$
    \\
    \hline\hline
      \multirow{4}{*}{\rotatebox{90}{Models}}
      &Perceive-IR~\cite{Perceive-IR10990319}& 30.93 & 0.902 & 32.99 & 0.948 & 40.05 & \underline{0.996} & 29.79 & 0.889 & 33.81 & 0.962 & 33.51 & 0.940
        \\
        &Defusion~\cite{DefusionLuo_2025_CVPR}& 30.66 & 0.905 & 33.33 & 0.952 & 40.21 & 0.995& 29.98 & 0.889 & 33.73& 0.962 & 33.58 & 0.940
        \\
        &AllRestorer~\cite{Allrestor11367271} &30.71 &0.903 & 33.17& 0.951 & 40.11 &0.991 & 29.63 &0.882& 33.62 & 0.959 & 33.45 & 0.937
        \\
        &BaryIR~\cite{baryir11417902} & 30.82 &0.907 & 33.05& 0.948& 40.20& 0.995 & 29.52& 0.881& 33.55 & 0.963 & 33.43 & 0.939
    \\
    \hline
 \multirow{6}{*}{\rotatebox{90}{Agents}}
&MAIR~\cite{MAIRjiang2026multi}
& 31.23 & 0.921 & 33.97 & 0.957 & 41.48 & \underline{0.996} & 34.08 & 0.965 & 34.02 & \textbf{0.971 }&34.96 &0.962
\\
&AgenticIR~\cite{agenticir}
& 31.10 & 0.920 & 34.17 & 0.961 & 41.25 & \underline{0.996} & 34.18 & 0.963 & 34.25 & \underline{0.969} &34.99  &0.962 
\\
&HybridAgent~\cite{hybridagentli2026hybrid}
& \underline{31.56} & \underline{0.923} & 34.03 & 0.959 & \underline{41.52} & \underline{0.996} & 34.29 & \underline{0.968} & \underline{34.28} & 0.968 & 35.17 &0.963
\\
&IAMAgent~\cite{wei2026iamagent}
& 31.55 & 0.922 & \underline{34.50} & \underline{0.962} & 41.44 & \underline{0.996} & \underline{34.31} &\textbf{0.974} & \textbf{34.40} & 0.965 &\underline{35.24}  &\underline{0.964}
\\
&Restore-R1~\cite{lu2026restorer1}
& 31.48 & 0.920 & 34.46 & \underline{0.962} & 41.39 & \textbf{0.997} & 34.15 & \underline{0.968} & 34.17 & 0.968 &35.13  &0.963
\\
&\cellcolor{gray!20}\textbf{RecurIR(Ours)}
&\cellcolor{gray!20}\textbf{31.93}
&\cellcolor{gray!20}\textbf{0.924}
&\cellcolor{gray!20}\textbf{34.79}
&\cellcolor{gray!20}\textbf{0.963}
&\cellcolor{gray!20}\textbf{41.96}
&\cellcolor{gray!20}\textbf{0.997}
&\cellcolor{gray!20}\textbf{34.81}
&\cellcolor{gray!20}\textbf{0.974}
&\cellcolor{gray!20}\textbf{34.40}
&\cellcolor{gray!20}\textbf{0.971}
&\cellcolor{gray!20}\textbf{35.58}
&\cellcolor{gray!20}\textbf{0.966}
\\
    \hline
\end{tabular}}
\end{table*}

\subsection{All-in-One Setting Results}

Table~\ref{tb:allrshbn} presents the quantitative results of restoration models and agent-based methods under the all-in-one R+S+H+B+N setting. Overall, agent-based methods achieve higher restoration quality across the evaluated tasks. Among the AiO-IR models, Defusion~\cite{DefusionLuo_2025_CVPR} achieves the highest average PSNR of 33.58 dB, whereas MAIR~\cite{MAIRjiang2026multi}, the lowest-performing agent in terms of average PSNR, achieves 34.96 dB. This 1.38 dB improvement highlights the advantages of adaptive tool coordination over a single unified restoration model in this evaluation setting. 

\textbf{Comparison with restoration models.}
Compared with existing all-in-one restoration models, RecurIR achieves an average PSNR of 35.58 dB, outperforming the strongest model-based baseline, Defusion~\cite{DefusionLuo_2025_CVPR}, by 2.00 dB. These results demonstrate the effectiveness of agentic restoration in coordinating specialized restoration operations to accommodate different degradation characteristics.

\textbf{Comparison with restoration agents.}
RecurIR also achieves the highest average PSNR and SSIM among the evaluated restoration agents. Compared with IAMAgent~\cite{wei2026iamagent}, the strongest competing method, RecurIR improves the average PSNR from 35.24 dB to 35.58 dB, corresponding to a gain of 0.34 dB. This improvement suggests that recursively discovering, challenging, and revising restoration decision rules can provide more effective guidance for restoration planning than conventional experience-driven decision strategies.

\textbf{Results on individual restoration tasks.}
RecurIR achieves the highest PSNR on deraining, desnowing, dehazing, and deblurring, reaching 31.93 dB, 34.79 dB, 41.96 dB, and 34.81 dB, respectively. For denoising, it obtains 34.40 dB, matching the best result achieved by IAMAgent~\cite{wei2026iamagent}. RecurIR also achieves competitive SSIM performance across all five tasks, indicating its effectiveness in handling diverse restoration requirements.

\begin{figure*} % use float package if you want it here
    \centerline{\includegraphics[width=1\linewidth]{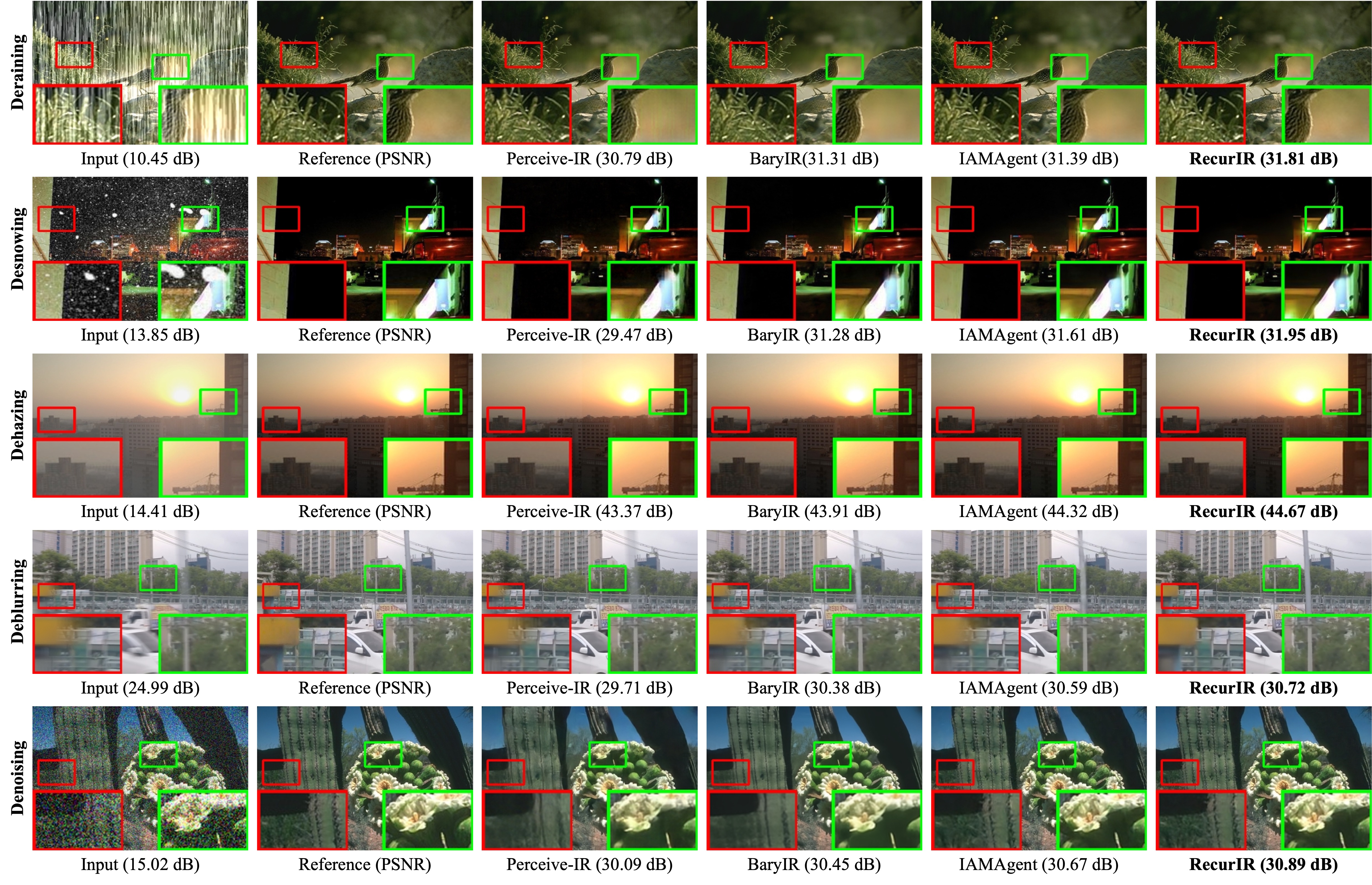}}
	\caption{Visual comparisons of different restoration methods. The red and green boxes highlight representative regions, with magnified views illustrating differences in artifact removal and detail preservation.}
 \label{fig:all-in-one}
\end{figure*}

Figure~\ref{fig:all-in-one} illustrates the visual comparisons across five restoration tasks. Compared with existing methods, RecurIR achieves more effective degradation removal while better preserving structural details and natural image appearance. As highlighted in the enlarged regions, competing methods exhibit varying degrees of residual degradation or detail loss.  In particular, RecurIR more effectively suppresses rain streaks and snow artifacts, recovers clearer building structures in hazy scenes, and preserves sharper object boundaries in blurred images. For denoising, it reduces noise while retaining finer textures in the restored regions. These improvements suggest
that adapting restoration decisions to degradation characteristics helps balance artifact removal and detail preservation.

\begin{table}
\centering
\caption{Quantitative comparison on three real-world benchmarks.}
\label{tb:allrealw}
\resizebox{\linewidth}{!}{
\begin{tabular}{c|ccc|ccc|cc}
    \hline
  \multirow{2}{*}{Methods} & \multicolumn{3}{c|}{RealRain-1k-L} & \multicolumn{3}{c|}{RTTS} & \multicolumn{2}{c}{SIDD}
    \\
    &PSNR $\uparrow$ &  SSIM $\uparrow$ &LPIPS $\downarrow$ &FADE $\downarrow$ &BRISQUE $\downarrow$ &NIMA $\uparrow$&PSNR $\uparrow$ & SSIM $\uparrow$
    \\
    \hline\hline
  ECFNet~\cite{gao2026emphasizing} & 23.69 & 0.756 & 0.401 & 1.394 & 29.105 &4.372 & 24.33 & 0.471 
    \\
      Perceive-IR~\cite{Perceive-IR10990319}& 27.31 & 0.901  & 0.372 & 1.264 & 23.694 &4.613 & 24.53 & 0.497
        \\
        MAIR~\cite{MAIRjiang2026multi}
&27.53 & 0.907 & \underline{0.335} & \underline{1.190} & \underline{23.231} & 4.874 & 24.68 & 0.500
\\
HybridAgent~\cite{hybridagentli2026hybrid}
&\underline{27.65} &0.909 &0.351 & 1.198 & 23.280 & 4.811 & \underline{24.70} & \underline{0.509}
\\
IAMAgent~\cite{wei2026iamagent}
&27.61 & \underline{0.912} & 0.344 & 1.203 & 23.284 & \underline{4.887} & 24.66 & 0.498
\\
\hline
      \cellcolor{gray!20}\textbf{RecurIR(Ours)} & \cellcolor{gray!20}\textbf{27.83} 
      &\cellcolor{gray!20}\textbf{0.915}
      & \cellcolor{gray!20}\textbf{0.329}
      & \cellcolor{gray!20}\textbf{1.158} 
      & \cellcolor{gray!20}\textbf{23.172}
      & \cellcolor{gray!20}\textbf{4.900} 
      & \cellcolor{gray!20}\textbf{24.74}
      &\cellcolor{gray!20}\textbf{0.521}
    \\
    \hline
\end{tabular}}
\end{table}

\begin{table*}
\centering
\caption{Quantitative comparison under individual and combined haze (H), rain (R), and noise (N) degradations.}
\label{tb:mix}
\resizebox{\linewidth}{!}{
\begin{tabular}{ccccccccccccccc}
    \hline
   Methods& H + R + N  & H + R &H + N& R + N & H &R&N
    \\
    \hline\hline
Perceive-IR~\cite{Perceive-IR10990319}&30.22/0.894	&31.55/0.945	&28.32/0.871	&30.52/0.849	&31.90/0.980	&38.95/0.983	&31.88/0.923	
\\
DSwinIR~\cite{DswinIR11304568} & 30.09/0.899 & 31.07/0.960 & 28.45/0.873 & 31.29/0.915 & 31.77/0.973 & 39.09/0.982 & 32.22/0.923 
\\
 MAIR~\cite{MAIRjiang2026multi}& 31.25/0.912 & 31.56/0.973 & 31.33/0.907 & 32.07/0.919 & \underline{39.43}/\textbf{0.993 }& \underline{40.17/0.992} &32.76/0.935
\\
AgenticIR~\cite{agenticir}& 31.31/0.912 & 31.62/0.971 & 31.44/0.910 & 32.15/\underline{0.921} & \underline{39.43/0.991} & 40.06/0.991 &\underline{32.83/0.937}
\\
HybridAgent~\cite{hybridagentli2026hybrid}& 31.22/0.912 & \underline{31.88}/\underline{0.976}& \underline{31.52/0.911} & \underline{32.31}/\textbf{0.922} & 39.39/\textbf{0.993} & 39.99/0.990 &32.54/0.933
\\
IAMAgent~\cite{wei2026iamagent}& \underline{31.52/0.926} & 31.70/0.975 & 31.09/0.905 & 31.98/0.920 & 39.21/0.990 & 40.11/\underline{0.992} &32.58/0.933
\\
Restore-R1~\cite{lu2026restorer1}& 31.20/0.910 & 31.73/\underline{0.976} & 31.27/0.910 & 32.22/0.918 & 39.08/\underline{0.991} & 39.96/0.989 &32.81/0.929
\\
\hline
\rowcolor{gray!20}\textbf{RecurIR(Ours)} &  \textbf{31.73/0.930} & \textbf{32.11/0.977} & \textbf{31.79/0.915} & \textbf{32.53}/\underline{0.921} & \textbf{39.48/0.993} & \textbf{40.28/0.994} &\textbf{33.09/0.942}

    \\
    \hline
\end{tabular}}
\end{table*}

\begin{figure*} % use float package if you want it here
    \centerline{\includegraphics[width=1\linewidth]{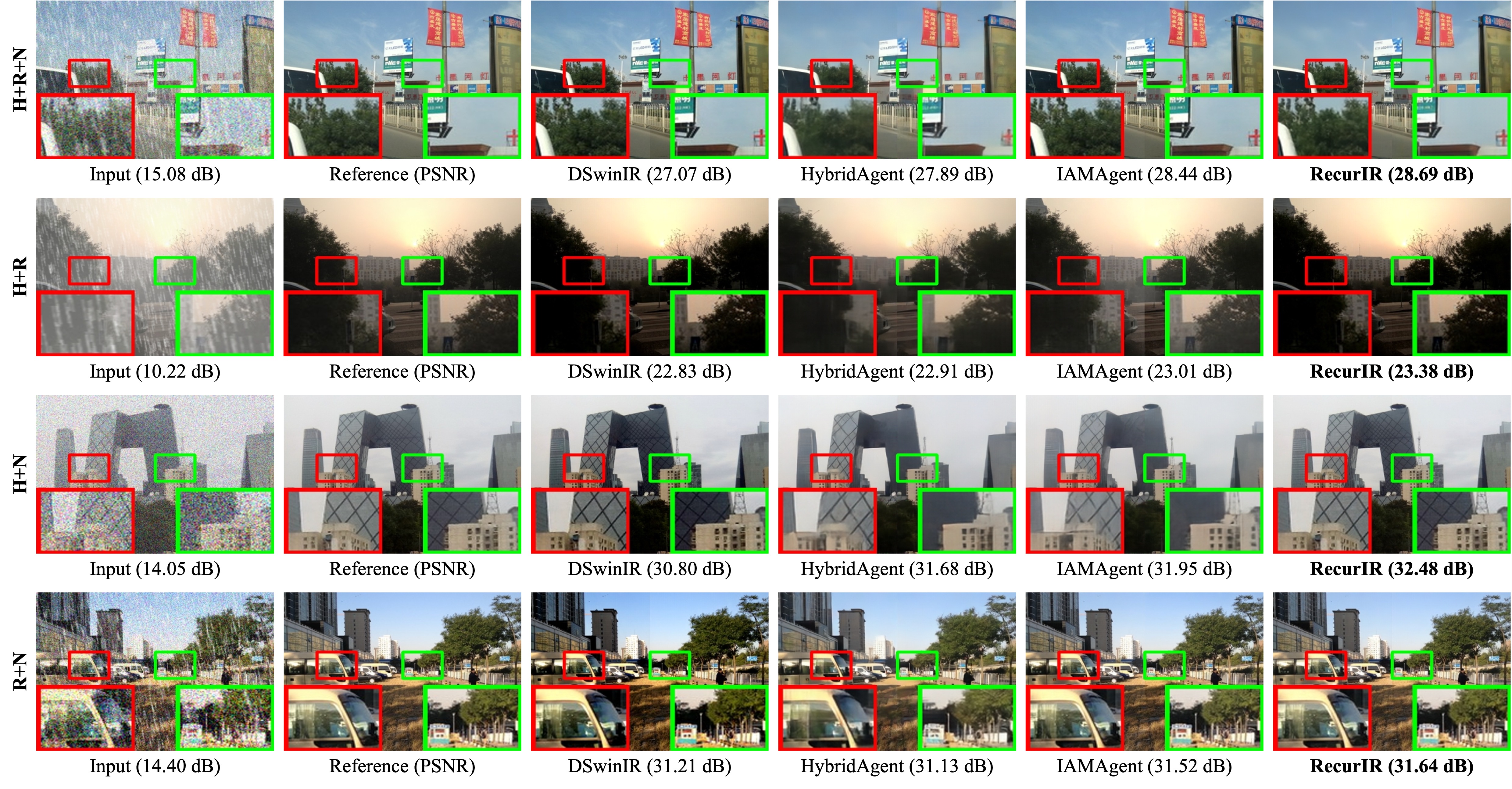}}
	\caption{Multi-degradation image restoration results across various degradation combinations.}
 \label{fig:mixm}
\end{figure*}

\subsection{Generalization}
\subsubsection{Evaluation on Real-world Scenes}
To assess the generalization ability of RecurIR, we evaluate different methods on three real-world benchmarks: RealRain-1k-L~\cite{realrain1li2022toward}, RTTS~\cite{RESIDEli2018benchmarking}, and SIDD~\cite{ssidabdelhamed2018high}. The quantitative comparisons are presented in Table~\ref{tb:allrealw}.
On RealRain-1k-L, RecurIR achieves the highest PSNR of 27.83 dB, outperforming the strongest competing method, HybridAgent~\cite{hybridagentli2026hybrid}, by 0.18 dB in PSNR. It also reduces LPIPS from 0.335, achieved by MAIR~\cite{MAIRjiang2026multi}, to 0.329, indicating improved perceptual quality. On RTTS, RecurIR consistently leads across all evaluation metrics, obtaining a FADE score of 1.158, a BRISQUE score of 23.172, and a NIMA score of 4.900. These results demonstrate its effectiveness in removing real-world haze while preserving visual quality.

\subsubsection{Generalization to Mixed Degradations}

We further investigate the restoration performance under mixed degradations involving haze, rain, and noise, considering both individual and combined degradation settings. Quantitative comparisons are presented in Table~\ref{tb:mix}, with visual results shown in Figure~\ref{fig:mixm}.

As shown in Table~\ref{tb:mix}, agent-based methods generally outperform model-based, particularly when multiple degradations coexist. While Perceive-IR~\cite{Perceive-IR10990319} and DSwinIR~\cite{DswinIR11304568} achieve competitive results on individual degradations, their performance is relatively limited under more complex degradation combinations. In comparison, restoration agents benefit from adaptive tool coordination and flexible execution strategies, leading to better restoration quality across different degradation settings.
Among the evaluated methods, RecurIR achieves the highest PSNR across all seven degradation settings. For the challenging H+R+N combination, RecurIR obtains 31.73 dB, exceeding the strongest model-based method, Perceive-IR~\cite{Perceive-IR10990319}, by 1.51 dB and the best competing agent, IAMAgent~\cite{wei2026iamagent}, by 0.21 dB. 
The qualitative comparisons in Figure~\ref{fig:mixm} further illustrate the restoration differences. AiO-IR models may leave residual haze, rain streaks, or noise, while also introducing over-smoothing and color distortions. Existing restoration agents alleviate some of these artifacts through sequential tool execution. RecurIR further improves structural recovery and visual consistency by adapting restoration decisions to changing degradation conditions. 

\subsection{Agent Efficiency and Preference Alignment}
\label{subsec:agent_efficiency}

Beyond restoration quality, an effective agent should minimize unnecessary operations while maintaining reliable performance and aligning with user preferences. We therefore evaluate rollback rate (RR), restoration success rate (SR), and preference alignment rate (PAR).
As shown in Table~\ref{tab:agent_efficiency}, RecurIR achieves the lowest RR (0.12) and the highest SR (0.94) and PAR (0.91) among the evaluated agents.  These improvements suggest that recursively discovering and revising decision rules enables more reliable restoration planning while better accommodating user requirements.

\begin{table}
\centering
\caption{Comparison of restoration agents in terms of rollback rate, restoration success rate, and preference alignment.}
\label{tab:agent_efficiency}
\begin{tabular}{cccc}
    \hline
    Method & RR $\downarrow$ & SR $\uparrow$ & PAR $\uparrow$ \\
    \hline\hline
    AgenticIR~\cite{agenticir}
    & 0.32 & 0.84 & \underline{0.88} \\
    MAIR~\cite{MAIRjiang2026multi}
    & 0.36 & \underline{0.90} & 0.80 \\
    HybridAgent~\cite{hybridagentli2026hybrid}
    & 0.38 & 0.88 & 0.82 \\
    IAMAgent~\cite{wei2026iamagent}
    & 0.32 & \underline{0.90} & 0.86 \\
    Restore-R1~\cite{lu2026restorer1}
    & \underline{0.28} & 0.88 & 0.86 \\
    \hline
    \rowcolor{gray!20}
    \textbf{RecurIR (Ours)}
    & \textbf{0.12} & \textbf{0.94} & \textbf{0.91} \\
    \hline
\end{tabular}
\end{table}

\subsection{Ablation Studies}

We conduct ablation studies to assess the contributions of three key components in RecurIR. As shown in Table~\ref{tab:abl1}, the baseline achieves 34.88 dB PSNR, with an RR of 0.35, an SR of 0.84, and a PAR of 0.79. Introducing PDM improves PSNR to 35.11 dB and reduces RR to 0.27, indicating that condition-dependent decision hypotheses help the agent identify more reliable restoration strategies. Incorporating RTEM further increases PSNR to 35.39 dB while reducing RR to 0.18. This improvement suggests that actively challenging learned decision rules provides informative training evidence beyond conventional restoration experiences.
With CPRM, the complete RecurIR achieves the best overall performance, reaching 35.58 dB PSNR, 0.12 RR, 0.94 SR, and 0.91 PAR. Compared with the baseline, it improves PSNR by 0.70 dB and reduces RR by 0.23. These results demonstrate the complementary benefits of discovering conditional decision rules, generating targeted challenges, and learning how to revise restoration policies under changing degradation conditions.

\begin{table}
    \centering
    \caption{Ablation study of the key components in RecurIR.}
    \label{tab:abl1}
    \resizebox{\linewidth}{!}{
    \begin{tabular}{cccccccc}
    \hline
    Method & PDM & RTEM & CPRM & RR $\downarrow$ & SR $\uparrow$& PAR $\uparrow$  & PSNR $\uparrow$ \\
    \hline\hline
    (a) & & & & 0.35 &0.84 &0.79 &  34.88 \\
    (b) & \ding{52} & & &0.27 &0.88 &0.86& 35.11 \\
    (c) & \ding{52} & \ding{52} &  &0.18 &0.92 &0.89& 35.39 \\
    (d) & \ding{52} & \ding{52} & \ding{52} & 0.12 & 0.94 & 0.91 & 35.58 
    \\
    \hline
    \end{tabular}}
\end{table}

\begin{figure}
    \centerline{\includegraphics[width=1\linewidth]{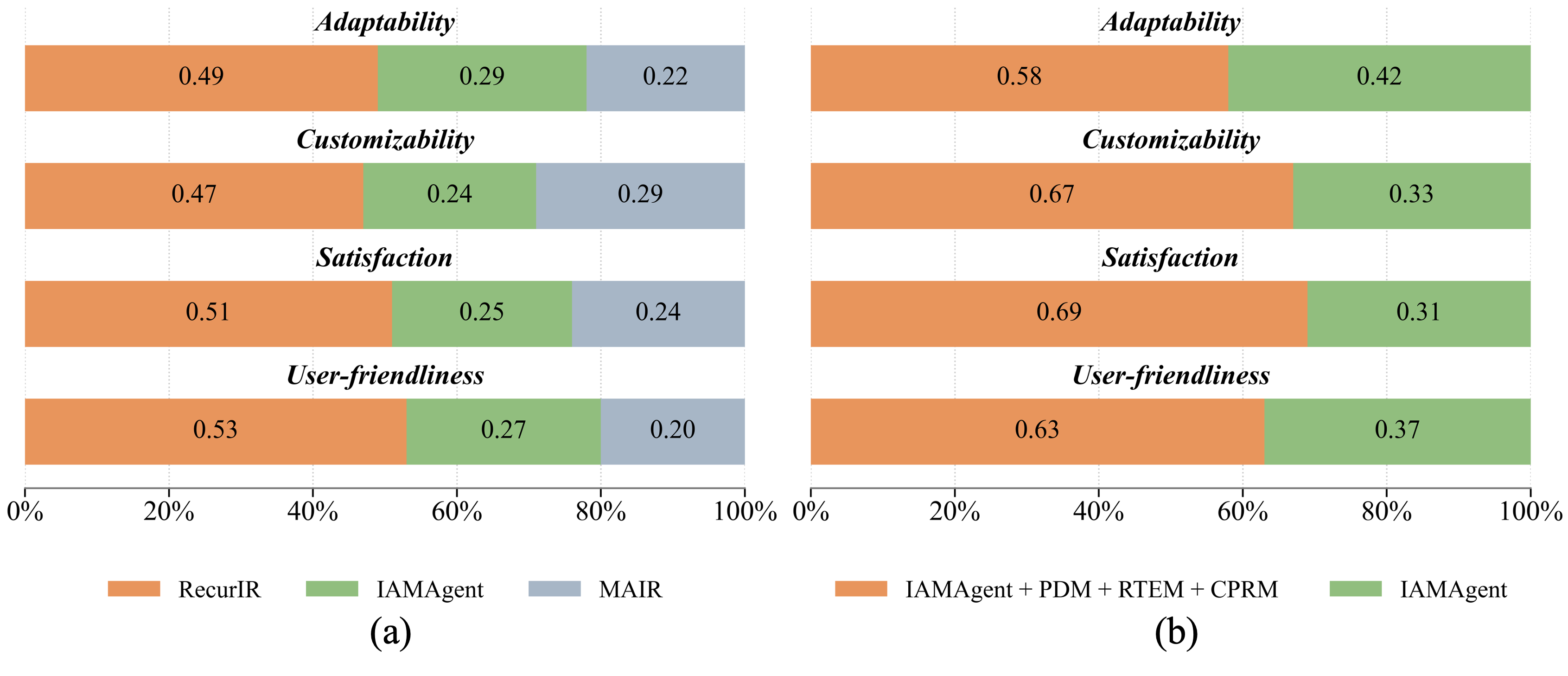}}
    \caption{User study results following the evaluation protocol of IAMAgent~\cite{wei2026iamagent}. (a) User preferences for RecurIR, IAMAgent~\cite{wei2026iamagent}, and MAIR~\cite{MAIRjiang2026multi} across four evaluation dimensions. (b) Comparison between the original IAMAgent and its enhanced version incorporating PDM, RTEM, and CPRM.}
    \label{fig:user_study}
\end{figure}

\subsection{User Study}

Following the evaluation protocol of IAMAgent~\cite{wei2026iamagent}, we conduct a user study involving 50 participants from diverse backgrounds to assess four aspects of restoration performance: adaptability, customizability, satisfaction, and user-friendliness. Satisfaction is evaluated through blind comparisons of anonymized restoration results, while the remaining dimensions are assessed through direct interaction with the systems to reflect users experiences with their responsiveness and controllability.

Figure~\ref{fig:user_study}(a) compares RecurIR with two representative restoration agents, IAMAgent and MAIR. RecurIR receives the highest preference ratios across all four dimensions. These results indicate that RecurIR provides a more favorable user experience, particularly in adapting restoration decisions to varying degradation conditions and producing satisfactory visual results.
To further examine the general applicability of our proposed mechanisms, we incorporate PDM, RTEM, and CPRM into IAMAgent~\cite{wei2026iamagent} while retaining its original restoration framework. As shown in Figure~\ref{fig:user_study}(b), the enhanced IAMAgent receives higher user preference across all four dimensions. These findings suggest that the proposed mechanisms can improve restoration adaptability and user-perceived quality beyond RecurIR itself.
\section{Conclusion}

In this paper, we propose RecurIR, a recursive self-improvement framework for learning condition-dependent restoration rule selection and revision. Through policy-hypothesis discovery, RecurIR identifies relationships between degradation conditions and restoration decisions. Rule-falsification-driven task evolution examines the applicability of learned rules, while change-conditioned policy revision learns when and how restoration strategies should be adjusted. Verified feedback progressively refines these condition-dependent relationships, enabling more reliable decisions as degradation conditions change. Extensive experiments demonstrate the effectiveness of RecurIR.
\section*{LLM Usage Statement}

In this work, we used generative AI tools to assist with manuscript editing and
scientific figure preparation. GPT-based tools were used to improve the
language, grammar, and readability of the manuscript without changing the
underlying scientific content.  The underlying data and numerical results in the figure were not generated or modified by AI.

All AI-assisted outputs were manually reviewed and verified by the authors.
The authors take full responsibility for the final manuscript, including all
text, figures, experimental results, scientific claims, and conclusions.

{
    \small
    \bibliographystyle{ieeenat_fullname}
    \bibliography{main}

@String(CVPR= {IEEE Conf. Comput. Vis. Pattern Recog.})

@String(ICCV= {Int. Conf. Comput. Vis.})

@String(ECCV= {Eur. Conf. Comput. Vis.})

@String(TIP  = {IEEE Trans. Image Process.})

@String(ICASSP=	{ICASSP})

@String(ICLR = {Int. Conf. Learn. Represent.})

@String(AAAI = {AAAI})

@String(CVPR  = {CVPR})

@String(ICCV  = {ICCV})

@String(ECCV  = {ECCV})

@String(TIP   = {IEEE TIP})

@String(TCSVT = {IEEE TCSVT})

@String(ICLR  = {ICLR})

@article{RESIDEli2018benchmarking,
  title={Benchmarking single-image dehazing and beyond},
  author={Li, Boyi and Ren, Wenqi and Fu, Dengpan and Tao, Dacheng and Feng, Dan and Zeng, Wenjun and Wang, Zhangyang},
  journal={TIP},
  volume={28},
  number={1},
  pages={492--505},
  year={2018},
  publisher={IEEE}
}

@inproceedings{Zamir2021Restormer,
    title={Restormer: Efficient Transformer for High-Resolution Image Restoration}, 
    author={Syed Waqas Zamir and Aditya Arora and Salman Khan and Munawar Hayat 
            and Fahad Shahbaz Khan and Ming-Hsuan Yang},
    booktitle={CVPR},
    year={2022}
}

@ARTICLE{DswinIR11304568,
  author={Wu, Gang and Jiang, Junjun and Jiang, Kui and Liu, Xianming and Nie, Liqiang},
  journal={IEEE Transactions on Pattern Analysis and Machine Intelligence}, 
  title={DSwinIR: Rethinking Window-Based Attention for Image Restoration}, 
  year={2026},
  volume={48},
  number={4},
  pages={4350-4366},
  }

@inproceedings{MLLMli2023blip,
  title={Blip-2: Bootstrapping language-image pre-training with frozen image encoders and large language models},
  author={Li, Junnan and Li, Dongxu and Savarese, Silvio and Hoi, Steven},
  booktitle={International conference on machine learning},
  pages={19730--19742},
  year={2023},
}

@InProceedings{MDTChen_2025_CVPR,
    author    = {Chen, Duosheng and Zhou, Shihao and Pan, Jinshan and Shi, Jinglei and Qu, Lishen and Yang, Jufeng},
    title     = {A Polarization-Aided Transformer for Image Deblurring via Motion Vector Decomposition},
    booktitle = {Proceedings of the Computer Vision and Pattern Recognition Conference (CVPR)},
    month     = {June},
    year      = {2025},
    pages     = {28061-28070}
}

@article{MLLMsingh2025openai,
  title={Openai gpt-5 system card},
  author={Singh, Aaditya and Fry, Adam and Perelman, Adam and Tart, Adam and Ganesh, Adi and El-Kishky, Ahmed and McLaughlin, Aidan and Low, Aiden and Ostrow, AJ and Ananthram, Akhila and others},
  journal={arXiv preprint arXiv:2601.03267},
  year={2025}
}

@article{MLLMliu2023visual,
  title={Visual instruction tuning},
  author={Liu, Haotian and Li, Chunyuan and Wu, Qingyang and Lee, Yong Jae},
  journal={Advances in neural information processing systems},
  volume={36},
  pages={34892--34916},
  year={2023}
}

@article{MAIRjiang2026multi,
  title={Multi-agent image restoration},
  author={Jiang, Xu and Li, Gehui and Chen, Bin and Zhang, Jian},
  journal={International Journal of Computer Vision},
  volume={134},
  number={5},
  pages={205},
  year={2026},
}

@article{LLMbrown2020language,
  title={Language models are few-shot learners},
  author={Brown, Tom and Mann, Benjamin and Ryder, Nick and Subbiah, Melanie and Kaplan, Jared D and Dhariwal, Prafulla and Neelakantan, Arvind and Shyam, Pranav and Sastry, Girish and Askell, Amanda and others},
  journal={Advances in neural information processing systems},
  volume={33},
  pages={1877--1901},
  year={2020}
}

@article{LLMtouvron2023llama,
  title={Llama 2: Open foundation and fine-tuned chat models},
  author={Touvron, Hugo and Martin, Louis and Stone, Kevin and Albert, Peter and Almahairi, Amjad and Babaei, Yasmine and Bashlykov, Nikolay and Batra, Soumya and Bhargava, Prajjwal and Bhosale, Shruti and others},
  journal={arXiv preprint arXiv:2307.09288},
  year={2023}
}

@article{LLMroziere2023code,
  title={Code llama: Open foundation models for code},
  author={Roziere, Baptiste and Gehring, Jonas and Gloeckle, Fabian and Sootla, Sten and Gat, Itai and Tan, Xiaoqing Ellen and Adi, Yossi and Liu, Jingyu and Sauvestre, Romain and Remez, Tal and others},
  journal={arXiv preprint arXiv:2308.12950},
  year={2023}
}

@inproceedings{lu2026restorer1,
  title={Restore-R1: Efficient Image Restoration Agents via Reinforcement Learning with Multimodal LLM Perceptual Feedback},
  author={Lu, Jianglin and Wu, Yuanwei and Zhao, Ziyi and Wang, Hongcheng and Jimenez, Felix and Majeedi, Abrar and Fu, Yun},
  booktitle={Proceedings of the IEEE/CVF Conference on Computer Vision and Pattern Recognition},
  pages={8629--8639},
  year={2026}
}

@inproceedings{hybridagentli2026hybrid,
  title={Hybrid agents for image restoration},
  author={Li, Bingchen and Li, Xin and Lu, Yiting and Chen, Zhibo},
  booktitle={Proceedings of the IEEE/CVF Conference on Computer Vision and Pattern Recognition},
  pages={22636--22647},
  year={2026}
}

@article{wei2026iamagent,
  title={IAMAgent: Towards An Interactive and Adaptive Multi-Agent System for Image Restoration},
  author={Wei, Yanyan and Zhang, Yilin and Zheng, Huan and Ren, Jiahuan and Xu, Xiaogang and Shi, Zenglin and Zhang, Zhao and Wang, Meng},
  journal={IEEE Transactions on Image Processing},
  year={2026},
  publisher={IEEE}
}

@inproceedings{agenticir,
  title={An intelligent agentic system for complex image restoration problems},
  author={Zhu, Kaiwen and Gu, Jinjin and You, Zhiyuan and Qiao, Yu and Dong, Chao},
  booktitle={International Conference on Learning Representations},
  volume={2025},
  pages={57985--58013},
  year={2025}
}

@inproceedings{RestoreAgent,
author = {Chen, Haoyu and Li, Wenbo and Gu, Jinjin and Ren, Jingjing and Chen, Sixiang and Ye, Tian and Pei, Renjing and Zhou, Kaiwen and Song, Fenglong and Zhu, Lei},
title = {RestoreAgent: autonomous image restoration agent via multimodal large language models},
year = {2024},
booktitle = {Proceedings of the 38th International Conference on Neural Information Processing Systems},
}

@ARTICLE{11342300,
  author={Feng, Hansen and Wang, Lizhi and Huang, Yiqi and Wang, Yuzhi and Zhu, Lin and Huang, Hua},
  journal={IEEE Transactions on Pattern Analysis and Machine Intelligence}, 
  title={Learning Physics-Informed Noise Models from Dark Frames for Low-Light Raw Image Denoising}, 
  year={2026},
  volume={48},
  number={4},
  pages={3952-3969},
 }

@article{DIDMDN,
  title={Density-Aware Single Image De-raining Using a Multi-stream Dense Network},
  author={He Zhang and Vishal M. Patel},
  journal={CVPR},
  year={2018},
  pages={695-704}
}

@article{Rain100,
  title={Deep Joint Rain Detection and Removal from a Single Image},
  author={Wenhan Yang and Robby T. Tan and Jiashi Feng and Jiaying Liu and Zongming Guo and Shuicheng Yan},
  journal={CVPR},
  year={2016},
  pages={1685-1694}
}

@article{Test100,
  title={Image De-Raining Using a Conditional Generative Adversarial Network},
  author={He Zhang and Vishwanath A. Sindagi and Vishal M. Patel},
  journal={TCSVT},
  year={2017},
  volume={30},
  pages={3943-3956}
}

@article{Gopro,
  title={Deep Multi-scale Convolutional Neural Network for Dynamic Scene Deblurring},
  author={Seungjun Nah and Tae Hyun Kim and Kyoung Mu Lee},
  journal={CVPR},
  year={2016},
  pages={257-265}
}

@ARTICLE{Allrestor11367271,
  author={Mao, Jiawei and Yang, Yu and Yin, Xuesong and Shao, Ling and Tang, Hao},
  journal={IEEE Transactions on Pattern Analysis and Machine Intelligence}, 
  title={All-in-One Transformer for Image Restoration Under Adverse Weather Degradations}, 
  year={2026},
  volume={48},
  number={6},
  pages={6628-6641},
 }

@ARTICLE{starir11429607,
  author={Cui, Yuning and Zamir, Syed Waqas and Yang, Ming-Hsuan and Knoll, Alois and Khan, Fahad Shahbaz and Khan, Salman},
  journal={IEEE Transactions on Pattern Analysis and Machine Intelligence}, 
  title={StarIR: Convolutional Image Restoration With Spatial-Frequency Fusion}, 
  year={2026},
  volume={},
  number={},
  pages={1-18},
  }

@ARTICLE{baryir11417902,
  author={Tang, Xiaole and He, Xiaoyi and Xu, Jiayi and Gu, Xiang and Sun, Jian},
  journal={IEEE Transactions on Pattern Analysis and Machine Intelligence}, 
  title={Learning Continuous Wasserstein Barycenter Space for Generalized All-in-One Image Restoration}, 
  year={2026},
  volume={},
  number={},
  pages={1-16},
  doi={10.1109/TPAMI.2026.3669121}}

@ARTICLE{pwfnet11433521,
  author={Jiang, Xingyu and Gao, Ning and Dou, Hongkun and Zhang, Xiuhui and Zhong, Xiaoqing and Deng, Yue and Li, Hongjue},
  journal={IEEE Transactions on Image Processing}, 
  title={Global Modeling Matters: A Fast, Lightweight, and Effective Baseline for Efficient Image Restoration}, 
  year={2026},
  volume={35},
  number={},
  pages={2740-2754},
 }

@inproceedings{ssidabdelhamed2018high,
  title={A high-quality denoising dataset for smartphone cameras},
  author={Abdelhamed, Abdelrahman and Lin, Stephen and Brown, Michael S},
  booktitle={Proceedings of the IEEE conference on computer vision and pattern recognition},
  pages={1692--1700},
  year={2018}
}

@article{realrain1li2022toward,
  title={Toward real-world single image deraining: A new benchmark and beyond},
  author={Li, Wei and Zhang, Qiming and Zhang, Jing and Huang, Zhen and Tian, Xinmei and Tao, Dacheng},
  journal={arXiv preprint arXiv:2206.05514},
  year={2022}
}

@inproceedings{SFNet,
  title={Selective Frequency Network for Image Restoration},
  author={Cui, Yuning and Tao, Yi and Bing, Zhenshan and Ren, Wenqi and Gao, Xinwei and Cao, Xiaochun and Huang, Kai and Knoll, Alois},
  booktitle={ICLR},
  year={2023}
}

@ARTICLE{FSNet,
  author={Cui, Yuning and Ren, Wenqi and Cao, Xiaochun and Knoll, Alois},
  journal={TPAMI}, 
  title={Image Restoration Via Frequency Selection}, 
  year={2023},
  volume={},
  number={},
  pages={1-16},
}

@ARTICLE{10558778,
  author={Rong, Jianxiang and Huang, Hua and Li, Jia},
  journal={IEEE Transactions on Image Processing}, 
  title={IMU-Assisted Accurate Blur Kernel Re-Estimation in Non-Uniform Camera Shake Deblurring}, 
  year={2024},
  volume={33},
  number={},
  pages={3823-3838},
 }

@ARTICLE{desnownet,
  author={Liu, Yun-Fu and Jaw, Da-Wei and Huang, Shih-Chia and Hwang, Jenq-Neng},
  journal={IEEE Transactions on Image Processing}, 
  title={DesnowNet: Context-Aware Deep Network for Snow Removal}, 
  year={2018},
  volume={27},
  number={6},
  pages={3064-3073},
  doi={10.1109/TIP.2018.2806202}
}

@INPROCEEDINGS{mspformer10095605,
  author={Chen, Sixiang and Ye, Tian and Liu, Yun and Liao, Taodong and Jiang, Jingxia and Chen, Erkang and Chen, Peng},
  booktitle={ICASSP 2023 - 2023 IEEE International Conference on Acoustics, Speech and Signal Processing (ICASSP)}, 
  title={MSP-Former: Multi-Scale Projection Transformer for Single Image Desnowing}, 
  year={2023},
  volume={},
  number={},
  pages={1-5},
  doi={10.1109/ICASSP49357.2023.10095605}}

@inproceedings{HDCW-Netchen2021all,
  title={All snow removed: Single image desnowing algorithm using hierarchical dual-tree complex wavelet representation and contradict channel loss},
  author={Chen, Wei-Ting and Fang, Hao-Yu and Hsieh, Cheng-Lin and Tsai, Cheng-Che and Chen, I and Ding, Jian-Jiun and Kuo, Sy-Yen and others},
  booktitle={Proceedings of the IEEE/CVF International Conference on Computer Vision},
  pages={4196--4205},
  year={2021}
}

@inproceedings{JSTASRchen2020jstasr,
  title={JSTASR: Joint size and transparency-aware snow removal algorithm based on modified partial convolution and veiling effect removal},
  author={Chen, Wei-Ting and Fang, Hao-Yu and Ding, Jian-Jiun and Tsai, Cheng-Che and Kuo, Sy-Yen},
  booktitle={Computer Vision--ECCV 2020: 16th European Conference, Glasgow, UK, August 23--28, 2020, Proceedings, Part XXI 16},
  pages={754--770},
  year={2020},
  organization={Springer}
}

@inproceedings{upid10.1145/3664647.3680560,
author = {Wen, Yuanbo and Gao, Tao and Chen, Ting},
title = {Unpaired Photo-realistic Image Deraining with Energy-informed Diffusion Model},
year = {2024},
booktitle = {Proceedings of the 32nd ACM International Conference on Multimedia},
pages = {360–369},
numpages = {10},
}

@InProceedings{liu2024xyscannet,
   title={XYScanNet: An Interpretable State Space Model for Perceptual Image Deblurring},
  author={Liu, Hanzhou and Liu, Chengkai and Xu, Jiacong and Jiang, Peng and Lu, Mi},
    booktitle = {Proceedings of the Computer Vision and Pattern Recognition Conference (CVPR)},
    year      = {2025},
    pages     = {779-789}
}

@inproceedings{autodir10.1007/978-3-031-73661-2_19,
author = {Jiang, Yitong and Zhang, Zhaoyang and Xue, Tianfan and Gu, Jinwei},
title = {AutoDIR: Automatic All-in-One Image Restoration with Latent Diffusion},
year = {2024},
booktitle = {Computer Vision – ECCV 2024: 18th European Conference, Milan, Italy, September 29–October 4, 2024, Proceedings, Part XL},
pages = {340–359},
numpages = {20},

}

@inproceedings{cui2025adair,
title={Ada{IR}: Adaptive All-in-One Image Restoration via Frequency Mining and Modulation},
author={Yuning Cui and Syed Waqas Zamir and Salman Khan and Alois Knoll and Mubarak Shah and Fahad Shahbaz Khan},
booktitle={The Thirteenth International Conference on Learning Representations},
year={2025}
}

@INPROCEEDINGS{diffunmix10656884,
  author={Zeng, Haijin and Cao, Jiezhang and Zhang, Kai and Chen, Yongyong and Luong, Hiep and Philips, Wilfried},
  booktitle={2024 IEEE/CVF Conference on Computer Vision and Pattern Recognition (CVPR)}, 
  title={Unmixing Diffusion for Self-Supervised Hyperspectral Image Denoising}, 
  year={2024},
  volume={},
  number={},
  pages={27820-27830},
}

@ARTICLE{CAPTNet10526271,
  author={Gao, Hu and Yang, Jing and Zhang, Ying and Wang, Ning and Yang, Jingfan and Dang, Depeng},
  journal={IEEE Transactions on Circuits and Systems for Video Technology}, 
  title={Prompt-Based Ingredient-Oriented All-in-One Image Restoration}, 
  year={2024},
  volume={34},
  number={10},
  pages={9458-9471},
  }

@inproceedings{ALGgao2024learning,
  title={Learning enriched features via selective state spaces model for efficient image deblurring},
  author={Gao, Hu and Ma, Bowen and Zhang, Ying and Yang, Jingfan and Yang, Jing and Dang, Depeng},
  booktitle={Proceedings of the 32nd ACM International Conference on Multimedia},
  pages={710--718},
  year={2024}
}

@inproceedings{LSSRgao2024learning,
  title={Learning optimal combination patterns for lightweight stereo image super-resolution},
  author={Gao, Hu and Yang, Jing and Zhang, Ying and Yang, Jingfan and Ma, Bowen and Dang, Depeng},
  booktitle={Proceedings of the 32nd ACM International Conference on Multimedia},
  pages={5566--5574},
  year={2024}
}

@INPROCEEDINGS{urban100,
  author={Huang, Jia-Bin and Singh, Abhishek and Ahuja, Narendra},
  booktitle={2015 IEEE Conference on Computer Vision and Pattern Recognition (CVPR)}, 
  title={Single image super-resolution from transformed self-exemplars}, 
  year={2015},
  volume={},
  number={},
  pages={5197-5206},
 }

@INPROCEEDINGS{BSD68,
  author={Martin, D. and Fowlkes, C. and Tal, D. and Malik, J.},
  booktitle={Proceedings Eighth IEEE International Conference on Computer Vision. ICCV 2001}, 
  title={A database of human segmented natural images and its application to evaluating segmentation algorithms and measuring ecological statistics}, 
  year={2001},
  volume={2},
  number={},
  pages={416-423 vol.2},
}

@article{efderainguo2025efficientderain+,
  title={EfficientDeRain+: Learning Uncertainty-Aware Filtering via RainMix Augmentation for High-Efficiency Deraining},
  author={Guo, Qing and Qi, Hua and Sun, Jingyang and Juefei-Xu, Felix and Ma, Lei and Lin, Di and Feng, Wei and Wang, Song},
  journal={International Journal of Computer Vision},
  volume={133},
  number={4},
  pages={2111--2135},
  year={2025},
}

@article{HIDE,
  title={Human-Aware Motion Deblurring},
  author={Ziyi Shen and Wenguan Wang and Xiankai Lu and Jianbing Shen and Haibin Ling and Tingfa Xu and Ling Shao},
  journal={2019 IEEE/CVF International Conference on Computer Vision (ICCV)},
  year={2019},
  pages={5571-5580}
}

@ARTICLE{10919160,
  author={Li, Yaowei and An, Hang and Zhang, Tong and Chen, Xiaoxuan and Jiang, Bo and Pan, Jinshan},
  journal={IEEE Transactions on Circuits and Systems for Video Technology}, 
  title={Omni-Deblurring: Capturing Omni-Range Context for Image Deblurring}, 
  year={2025},
  volume={},
  number={},
  pages={1-1},
  }

@article{60talebi2018nima,
  title={NIMA: Neural image assessment},
  author={Talebi, Hossein and Milanfar, Peyman},
  journal={IEEE transactions on image processing},
  volume={27},
  number={8},
  pages={3998--4011},
  year={2018},
  publisher={IEEE}
}

@inproceedings{59mittal2011blind,
  title={Blind/referenceless image spatial quality evaluator},
  author={Mittal, Anish and Moorthy, Anush K and Bovik, Alan C},
  booktitle={2011 conference record of the forty fifth asilomar conference on signals, systems and computers (ASILOMAR)},
  pages={723--727},
  year={2011},
  organization={IEEE}
}

@article{58choi2015referenceless,
  title={Referenceless prediction of perceptual fog density and perceptual image defogging},
  author={Choi, Lark Kwon and You, Jaehee and Bovik, Alan Conrad},
  journal={IEEE Transactions on Image Processing},
  volume={24},
  number={11},
  pages={3888--3901},
  year={2015},
  publisher={IEEE}
}

@inproceedings{54zhang2018unreasonable,
  title={The unreasonable effectiveness of deep features as a perceptual metric},
  author={Zhang, Richard and Isola, Phillip and Efros, Alexei A and Shechtman, Eli and Wang, Oliver},
  booktitle={Proceedings of the IEEE conference on computer vision and pattern recognition},
  pages={586--595},
  year={2018}
}

@article{gao2026emphasizing,
  title={Emphasizing crucial features for efficient image restoration},
  author={Gao, Hu and Ma, Bowen and Zhang, Ying and Yang, Jingfan and Yang, Jing and Wang, Xingjian and Dang, Depeng},
  journal={Pattern Recognition},
  pages={113575},
  year={2026},
  publisher={Elsevier}
}

@inproceedings{pptformerwang2025intra,
  title={Intra and Inter Parser-Prompted Transformers for Effective Image Restoration},
  author={Wang, Cong and Pan, Jinshan and Wang, Liyan and Wang, Wei},
  booktitle={Proceedings of the AAAI Conference on Artificial Intelligence},
  volume={39},
  number={7},
  pages={7609--7618},
  year={2025}
}

@article{gao2025mixed,
  title={Mixed hierarchy network for image restoration},
  author={Gao, Hu and Zhang, Ying and Yang, Jing and Dang, Depeng},
  journal={Pattern Recognition},
  volume={161},
  pages={111313},
  year={2025},
}

@ARTICLE{Perceive-IR10990319,
  author={Zhang, Xu and Ma, Jiaqi and Wang, Guoli and Zhang, Qian and Zhang, Huan and Zhang, Lefei},
  journal={IEEE Transactions on Image Processing}, 
  title={Perceive-IR: Learning to Perceive Degradation Better for All-in-One Image Restoration}, 
  year={2025},
  volume={},
  number={},
  pages={1-1},
  }

@article{deanetchen2024dea,
  title={DEA-Net: Single image dehazing based on detail-enhanced convolution and content-guided attention},
  author={Chen, Zixuan and He, Zewei and Lu, Zhe-Ming},
  journal={IEEE Transactions on Image Processing},
  year={2024},
  publisher={IEEE}
}

@article{FDTANetgao2025frequency,
  title={Frequency domain task-adaptive network for restoring images with combined degradations},
  author={Gao, Hu and Ma, Bowen and Zhang, Ying and Yang, Jingfan and Yang, Jing and Dang, Depeng},
  journal={Pattern Recognition},
  volume={158},
  pages={111057},
  year={2025},
  publisher={Elsevier}
}

@inproceedings{PGH2Netisu2025prior,
  title={Prior-guided hierarchical harmonization network for efficient image dehazing},
  author={Su, Xiongfei and Li, Siyuan and Cui, Yuning and Cao, Miao and Zhang, Yulun and Chen, Zheng and Wu, Zongliang and Wang, Zedong and Zhang, Yuanlong and Yuan, Xin},
  booktitle={Proceedings of the AAAI Conference on Artificial Intelligence},
  volume={39},
  number={7},
  pages={7042--7050},
  year={2025}
}

@InProceedings{DefusionLuo_2025_CVPR,
    author    = {Luo, Wenyang and Qin, Haina and Chen, Zewen and Wang, Libin and Zheng, Dandan and Li, Yuming and Liu, Yufan and Li, Bing and Hu, Weiming},
    title     = {Visual-Instructed Degradation Diffusion for All-in-One Image Restoration},
    booktitle = {Proceedings of the IEEE/CVF Conference on Computer Vision and Pattern Recognition (CVPR)},
    month     = {June},
    year      = {2025},
    pages     = {12764-12777}
}

@InProceedings{VLUNetZeng_2025_CVPR,
    author    = {Zeng, Haijin and Wang, Xiangming and Chen, Yongyong and Su, Jingyong and Liu, Jie},
    title     = {Vision-Language Gradient Descent-driven All-in-One Deep Unfolding Networks},
    booktitle = {Proceedings of the IEEE/CVF Conference on Computer Vision and Pattern Recognition (CVPR)},
    month     = {June},
    year      = {2025},
    pages     = {7524-7533}
}

@ARTICLE{PEUNet10830558,
  author={Guo, Xin and Wang, Xi and Fu, Xueyang and Zha, Zheng-Jun},
  journal={IEEE Transactions on Circuits and Systems for Video Technology}, 
  title={Deep Unfolding Network for Image Desnowing With Snow Shape Prior}, 
  year={2025},
  volume={35},
  number={5},
  pages={4740-4752},
 }

@article{gao2026causal,
  title={Causal-AgentIR: Self-Evolving Causal Memory for Adaptive Image Restoration Agents},
  author={Gao, Hu and Chen, Yulong and Ma, Lizhuang},
  journal={arXiv preprint arXiv:2607.21125},
  year={2026}
}

@INPROCEEDINGS{cui2026sear,
  author        = {Cui, Shuang and Ji, Fan and Sun, Guanglong and Guo, Yufei and Tang, Xiongxin and Li, Jiangmeng and Xu, Fanjiang},
  title         = {Self-Evolving Agentic Image Restoration via Deliberate Planning and Intuitive Execution},
  year          = {2026},
  pages={392--411},
 booktitle={Proceedings of the European Conference on Computer Vision},
}

@INPROCEEDINGS{zhang2026tir,
  author        = {Zhang, Yisheng and Jia, Guoli and Hu, Haote and Zhao, Shanxu and Zhao, Kaikai and Sun, Long and Long, Xinwei and Tian, Kai and Jiang, Che and Liu, Zhaoxiang and Wang, Kai and Lian, Shiguo and Zhang, Kaiyan and Zhou, Bowen},
  title         = {{TIR-Agent}: Training an Explorative and Efficient Agent for Image Restoration},
  year          = {2026},
  booktitle={Proceedings of the European Conference on Computer Vision},
  pages={ 523--540},

}

@article{zhuang2026evoir,
  title={EvoIR-Agent: Self-Evolving Image Restoration Agentic System via Experience-Driven Learning},
  author={Zhuang, Kailin and Wu, Jiawei and Jin, Zhi},
  journal={arXiv preprint arXiv:2605.22208},
  year={2026}
}

@article{shao2026skillir,
  title={SkillIR: Evolving Scene-Aware Skills for Agentic Image Restoration},
  author={Shao, Jie and Hu, Shengkai and Zhang, Xu and Song, Beihang and Jing, Yongcheng and Wu, Xu and Wan, Jun},
  journal={arXiv preprint arXiv:2609.21468},
  year={2026}
}

@article{zhang2026preact,
  title={Look Before You Restore: Learning to Predict Before Acting for Agentic Image Restoration},
  author={Zhang, Xu and Wan, Jun and Zhang, Lefei and Zhang, Liangpei},
  journal={arXiv preprint arXiv:2610.20384},
  year={2026}
}

@article{kodak,
  title={Kodak lossless true color image suite},
  author={Franzen, Rich},
  journal={source: http://r0k. us/graphics/kodak},
  volume={4},
  number={2},
  pages={9},
  year={1999}
}
}

% WARNING: do not forget to delete the supplementary pages from your submission 
\clearpage
\setcounter{page}{1}
\maketitlesupplementary

\section{Motivation Analysis}
\label{sup:motivation}

Existing agent-based image restoration methods improve restoration decisions through adaptive planning, experience accumulation, and execution feedback. By reusing successful restoration trajectories or refining previously acquired strategies, these agents can effectively coordinate restoration tools under different degradation conditions. However, a successful restoration strategy under one condition does not necessarily remain effective when degradation severity, spatial distribution, or interactions change. For example, an agent may learn from previous executions that denoising before deblurring produces favorable results. Nevertheless, this experience alone does not explicitly indicate whether the same operation order remains preferable under stronger noise, how denoising strength should vary across different regions, or which operations should be modified when multiple degradations overlap spatially. Although existing agents can adapt their decisions through planning and feedback, the conditions governing the applicability of learned restoration rules are not always explicitly characterized.

As summarized in Table~\ref{tab:memory_comparison0}, the key motivation of RecurIR is to move beyond reusing successful restoration experiences toward explicitly learning the relationships between degradation conditions and restoration decisions. The central challenge is therefore not simply identifying an effective restoration strategy, but determining the conditions under which that strategy should be selected, retained, or revised. This requires the agent to explicitly characterize how degradation variations affect the relative effectiveness of competing restoration decisions and to identify the boundaries within which previously learned preferences remain reliable.

Motivated by this observation, RecurIR introduces a recursive learning process that discovers condition-dependent decision hypotheses, actively examines their applicability boundaries, and learns corresponding restoration policy revisions. Through verified feedback, the acquired knowledge is progressively refined to support more reliable restoration decisions under changing degradation conditions.

\begin{table*}[t]
\centering
\caption{Comparison between experience-driven restoration agents
and RecurIR in restoration knowledge learning and decision-making.}
\label{tab:memory_comparison0}

\renewcommand{\arraystretch}{1.15}
\setlength{\tabcolsep}{7pt}

\begin{tabularx}{\textwidth}{
>{\raggedright\arraybackslash}p{0.18\textwidth}
>{\raggedright\arraybackslash}X
>{\raggedright\arraybackslash}X
}
\toprule
\textbf{Aspect}
& \textbf{Existing Experience-Driven Agents}
& \textbf{RecurIR (Ours)}
\\
\midrule
\hline

\textbf{Learned Knowledge}
& Historical restoration experiences and effective strategies
& \textcolor{blue!75!black}{Condition-dependent relationships between degradation characteristics and restoration rules}
\\
\hline
\textbf{Memory Content}
& Previously executed operations and their restoration outcomes
& \textcolor{blue!75!black}{Conditions governing when restoration rules should be selected, retained, or revised}
\\
\hline
\textbf{Decision Basis}
& Experience retrieval, restoration planning, and execution feedback
& \textcolor{blue!75!black}{Conditional rule matching, applicability assessment, and policy revision}
\\
\hline
\textbf{Condition Changes}
& Replanning based on previously acquired experiences
& \textcolor{blue!75!black}{Learning and updating rule-selection relationships as degradation conditions change}
\\
\hline
\textbf{Learning Objective}
& Improving restoration decisions through accumulated experiences
& \textcolor{blue!75!black}{Learning when and how to select and revise restoration rules under different degradation conditions}
\\

\bottomrule
\end{tabularx}
\end{table*}

\section{Experimental Setup}
\label{sec:exset}
We evaluate our method under both all-in-one and task-specific settings.

\subsection{Datasets}
\textbf{Task-specific setting.} 
We evaluate the framework separately on image denoising, deblurring, deraining, dehazing, and desnowing. Unlike conventional task-specific methods that are explicitly provided with the degradation category, RecurIR receives only the degraded image and a generic restoration instruction unless otherwise specified. The degradation agent first estimates the degradation condition, after which the planner selects an appropriate restoration tool or tool sequence from the shared library.
\textbf{i)} Image deraining: we evaluate on several benchmark datasets, including Rain100H~\cite{Rain100}, Rain100L~\cite{Rain100}, Test100~\cite{Test100}, and Test1200~\cite{DIDMDN}.
\textbf{ii)} Image desnowing: we utilize the Snow100K~\cite{desnownet}, SRRS~\cite{JSTASRchen2020jstasr}, and CSD~\cite{HDCW-Netchen2021all} datasets. Following the protocol in~\cite{FSNet}, we randomly sample 2,000 images for evaluation.
\textbf{iii)} Image dehazing: we evaluate RecurIR on the daytime synthetic subsets of RESIDE~\cite{RESIDEli2018benchmarking}. 
\textbf{iv)} Image deblurring: we evaluate RecurIR on the GoPro dataset~\cite{Gopro} and HIDE dataset~\cite{HIDE}.
\textbf{v)} Image denoising: Evaluation is performed on CBSD68~\cite{BSD68}, Urban100~\cite{urban100}, and Kodak24~\cite{kodak}. Additive white Gaussian noise is generated with $\sigma \in {15,25,50}$.

\textbf{All-in-one setting.}
We evaluate on Test100~\cite{Test100} for deraining, Snow100K~\cite{desnownet} for desnowing, SOTS-Indoor~\cite{RESIDEli2018benchmarking} for dehazing, GoPro~\cite{Gopro} for deblurring, and CBSD68~\cite{BSD68} for denoising.

\subsection{Evaluation Metrics}

For paired datasets, we report peak signal-to-noise ratio (PSNR), structural similarity index measure (SSIM), and learned perceptual image patch similarity (LPIPS)~\cite{54zhang2018unreasonable}. 
For real-world images without reference images, we report Blind/Referenceless Image Spatial Quality Evaluator (BRISQUE)~\cite{59mittal2011blind},  Fog Aware Density Evaluator (FADE)~\cite{58choi2015referenceless}, and Neural Image Assessment (NIMA)~\cite{60talebi2018nima}. For PSNR, SSIM, and NIMA, higher values indicate better performance, whereas lower values are preferred for LPIPS, FADE, and BRISQUE. In all tables, the best and second-best results are highlighted in bold and underlined, respectively.

Beyond restoration quality, an effective agent should minimize unnecessary operations while maintaining reliable performance and aligning with user preferences. We therefore evaluate rollback rate (RR), restoration success rate (SR), and preference alignment rate (PAR). 

\textbf{Rollback rate.}
RR measures the proportion of executed restoration operations that are subsequently reverted:
\begin{equation}
\mathrm{RR}=
\frac{N_{\mathrm{rollback}}}{N_{\mathrm{execution}}},
\label{eq:rollback_rate}
\end{equation}
where $N_{\mathrm{rollback}}$ and $N_{\mathrm{execution}}$ denote the numbers of reverted and executed operations, respectively. A lower RR indicates fewer unsuccessful operations requiring rollback.

\textbf{Restoration success rate.}
SR evaluates the proportion of restored images that improve upon their inputs without introducing severe artifacts:
\begin{equation}
\mathrm{SR}=\frac{1}{N}
\sum_{i=1}^{N}
\mathbb{I}
\left[
Q(I_i^{\mathrm{out}})
>
Q(I_i^{\mathrm{in}})
\land
A(I_i^{\mathrm{out}})=0
\right],
\label{eq:success_rate}
\end{equation}
where $Q(\cdot)$ is a composite restoration quality score, $A(\cdot)$ detects severe artifacts, and $\mathbb{I}[\cdot]$ is the indicator function.

\textbf{Preference alignment rate.}
To assess preference alignment, participants evaluate anonymized restoration results under instructions such as ``preserve natural details'', ``prefer stronger artifact removal'', and ``avoid over-enhancement''. PAR is calculated as:
\begin{equation}
\mathrm{PAR}=
\frac{N_{\mathrm{agree}}}{N_{\mathrm{total}}},
\label{eq:preference_alignment}
\end{equation}
where $N_{\mathrm{agree}}$ denotes the number of comparisons consistent with participant preferences, and $N_{\mathrm{total}}$ is the total number of comparisons. Each comparison involves at least 50 independent participants.

\begin{table}
\caption{Restoration tools used in the shared tool library.}
    \label{tab:restools}
    \centering
    \begin{tabular}{c|c}
    \hline
         Task& Tools
         \\
         \hline
         \hline
        Derain&  PPTformer~\cite{pptformerwang2025intra}, EfDeRain+~\cite{efderainguo2025efficientderain+}, MHNet~\cite{gao2025mixed}
        \\
        Desnow & MSP-Former~\cite{mspformer10095605}, PW-FNet~\cite{pwfnet11433521},PEUNet~\cite{PEUNet10830558}
        \\
        Dehaze & PGH$^2$Net~\cite{PGH2Netisu2025prior},  DEA-Net-CR~\cite{deanetchen2024dea}, ECFNet~\cite{gao2026emphasizing}
        \\
        Deblur &  Omni-Deblurring~\cite{10919160},ALGNet~\cite{ALGgao2024learning}, MDT~\cite{MDTChen_2025_CVPR} 
        \\
        Denoise & FSNet~\cite{FSNet}, StarIR~\cite{starir11429607},  Restormer~\cite{Zamir2021Restormer}
        \\
        All-in-one & Perceive-IR~\cite{Perceive-IR10990319}, BaryIR~\cite{baryir11417902}, Defusion~\cite{DefusionLuo_2025_CVPR}
         \\
         \hline
    \end{tabular}
\end{table}

\subsection{Training Details}

We follow the implementation settings of MAIR~\cite{MAIRjiang2026multi}.  The restoration tool library is listed in Table~\ref{tab:restools}. We use publicly available pretrained checkpoints whenever possible; otherwise, the tools are retrained following their original implementation settings. All restoration tools remain frozen during agent training. For fair comparison, all agent-based methods use the same restoration tool library.

\begin{table}[t]
\centering
\caption{Initial hyperparameter settings of RecurIR.}
\label{tab:hyperparameters}
\begin{tabular}{ccc}
\hline
Module & Parameter  & Initial Value  \\
\hline\hline
\multirow{2}{*}{PDM}
& $\tau$  & 0.5
\\
& $\lambda_m$  & 0.5
 \\
\hline
\multirow{5}{*}{RTEM}
& $\tau_m$  & 0.10
\\
& $\epsilon$  & 0.10
 \\
& $\tau_b$  & 0.2
\\
& $\lambda_b$  & 0.5
 \\
& $\lambda_c$ & 0.1
\\
\hline
\multirow{3}{*}{CPRM}
& $\lambda_p$  & 1.0
 \\
& $\lambda_r$  & 0.1
 \\
& $\tau_g$  & 0.5
 \\
\hline
\multirow{2}{*}{Verification}
& $\delta_{\mathrm{tr}}$  & 0.05
 \\
& $\epsilon_{\mathrm{ret}}$  & 0.05
 \\
\hline
\end{tabular}
\end{table}

\subsection{Hyperparameter Settings}

Table~\ref{tab:hyperparameters} summarizes the main hyperparameters of RecurIR and their initial values. In PDM, $\tau$ controls the temperature of preference prediction, while $\lambda_m$ balances preference classification and relative benefit regression. For RTEM, $\tau_m$ and $\epsilon$ determine the confidence margin and minimum quality improvement required for rule falsification, respectively. The parameter $\tau_b$ controls the sensitivity of boundary exploration, while $\lambda_b$ and $\lambda_c$ balance boundary exploration and task generation costs.
For CPRM, $\lambda_p$ and $\lambda_r$ control policy revision supervision and retention regularization, respectively, while $\tau_g$ determines whether a predicted revision is applied. The threshold $\epsilon$ is also used to identify meaningful improvements when constructing revision targets. Finally, $\delta_{\mathrm{tr}}$ and $\epsilon_{\mathrm{ret}}$ specify the minimum transfer improvement and maximum allowable retention degradation for accepting policy updates. 

\section{More Experiments}
\label{sec:moreex}

\begin{table*}
\centering
\caption{Image deraining results in the task-specific setting.}
\label{tb:derain}
    \resizebox{\linewidth}{!}{
\begin{tabular}{c|cccccccc|cc}
    \hline
    \multicolumn{1}{c|}{} & \multicolumn{2}{c}{Test100}  & \multicolumn{2}{c}{Test1200} & \multicolumn{2}{c}{Rain100H} & \multicolumn{2}{c|}{Rain100L} & \multicolumn{2}{c}{Average} 
    \\
   Methods &PSNR $\uparrow$ &  SSIM $\uparrow$  & PSNR $\uparrow$ & SSIM $\uparrow$ &PSNR $\uparrow$ &SSIM $\uparrow$ & PSNR $\uparrow$&SSIM $\uparrow$ &PSNR $\uparrow$ & SSIM $\uparrow$
    \\
    \hline\hline
     EfDeRain+~\cite{efderainguo2025efficientderain+} &31.10 &0.911 &33.12 & 0.925 & \textbf{34.57} &\textbf{0.957} & 39.03 & 0.972 &34.46 & 0.941
     \\
     MAIR~\cite{MAIRjiang2026multi}
& 31.23 & 0.921 & 33.28 & 0.927 & 34.40 & 0.955 & 39.99 & \underline{0.984} & 34.73 & 0.947
\\
AgenticIR~\cite{agenticir}
& 31.10 & 0.920 & 33.31 & \underline{0.928} & 34.52 & \underline{0.956} & 39.71 & 0.980 & 34.66 & 0.946
\\
HybridAgent~\cite{hybridagentli2026hybrid}
& \underline{31.56} & \underline{0.923} &  33.39 &0.926 & 34.28 &\underline{0.956} & 39.72 & 0.981 & 34.74 & \underline{0.947}
\\
IAMAgent~\cite{wei2026iamagent}
& 31.55 & 0.922 & \underline{33.43} & 0.925 & 34.55 & 0.951 & 39.41 & 0.979 & 34.74 & 0.944
\\
Restore-R1~\cite{lu2026restorer1}
& 31.48 & 0.920 & 33.25 & 0.926 & 34.27 & 0.955 & \underline{40.00} & 0.983 & \underline{34.75} &0.946
\\
      \hline
    \rowcolor{gray!20}  \textbf{RecurIR(Ours)}  & \textbf{31.93}	&\textbf{0.924}	&\textbf{33.61}	&\textbf{0.930}	&\underline{34.56}	&\textbf{0.957}	&\textbf{40.13}	&\textbf{0.985}	&\textbf{35.06}	&\textbf{0.949}
    \\
    \hline
\end{tabular}}
\end{table*}

\subsection{Task-specific Setting Results}

We further evaluate RecurIR in task-specific settings, covering five representative restoration tasks: image deraining, desnowing, dehazing, deblurring, and denoising.

\subsubsection{Image Deraining}

Following the evaluation protocol in~\cite{gao2025mixed}, we compute PSNR and SSIM on the Y channel of the YCbCr color space. Table~\ref{tb:derain} presents the results on four deraining benchmarks.
RecurIR achieves the highest reported average PSNR of 35.06 dB, outperforming Restore-R1~\cite{lu2026restorer1}, the strongest competing agent in average PSNR, by 0.31 dB. Compared with the task-specific EfDeRain+~\cite{efderainguo2025efficientderain+}, RecurIR improves average PSNR by 0.60 dB. Across individual datasets, RecurIR obtains the best PSNR on Test100, Test1200, and Rain100L. On Rain100H, it achieves 34.56 dB, only 0.01 dB below EfDeRain+. These results indicate that RecurIR maintains competitive performance across different rain degradation conditions.

\begin{table}
    \centering
        \caption{Image  desnowing results in the task-specific setting.}
    \label{tab:snow}
    \resizebox{\linewidth}{!}{
    \begin{tabular}{c|cccccc}
    \hline
    \multicolumn{1}{c|}{} & \multicolumn{2}{c}{CSD}  & \multicolumn{2}{c}{SRRS} & \multicolumn{2}{c}{Snow100K}
    \\
   Methods & PSNR $\uparrow$ & SSIM $\uparrow$  & PSNR $\uparrow$ & SSIM $\uparrow$ & PSNR $\uparrow$ & SSIM $\uparrow$
   \\
   \hline
   \hline
PEUNet~\cite{PEUNet10830558}&37.28 &0.97 &31.89 &\textbf{0.98} &34.11 &\textbf{0.96}
\\
PW-FNet~\cite{pwfnet11433521}&-&-&32.74&\textbf{0.98}&\underline{34.50}&\underline{0.95}
\\
 StarIR~\cite{starir11429607}  & \underline{38.57} & \textbf{0.99} & \underline{32.75} & \textbf{0.98} & 34.46 & \textbf{0.96}
\\
 MAIR~\cite{MAIRjiang2026multi}
&38.41  &\underline{0.98}  &32.70 &\underline{0.97}   & 33.97 &\textbf{0.96}
\\
AgenticIR~\cite{agenticir}
&38.55  &\underline{0.98}  &32.73 &\underline{0.97} & 34.17 & \textbf{0.96}
\\
HybridAgent~\cite{hybridagentli2026hybrid}
&38.29  &0.97  &32.68 &0.96 & 34.03 & \textbf{0.96}
\\
IAMAgent~\cite{wei2026iamagent}
&38.38  &\underline{0.98}  &32.74 &\textbf{0.98} & \underline{34.50} & \textbf{0.96}
\\
Restore-R1~\cite{lu2026restorer1}
&38.49  &0.97  &32.66 &\underline{0.97}  &34.46  &\textbf{0.96}
\\
\hline
\rowcolor{gray!20} \textbf{RecurIR(Ours)} &\textbf{38.59}	&\textbf{0.99}  &\textbf{32.93}	&\textbf{0.98}		&\textbf{34.79}&\textbf{0.96}

         \\
         \hline
    \end{tabular}}
\end{table}

\begin{table}
    \centering
        \caption{Image dehazing results in the task-specific setting.}
    \label{tab:sot}
    \resizebox{\linewidth}{!}{
    \begin{tabular}{c|cccc}
    \hline
    \multicolumn{1}{c|}{} & \multicolumn{2}{c}{SOTS-Indoor}  & \multicolumn{2}{c}{SOTS-Outdoor} 
    \\
   Methods & PSNR $\uparrow$ & SSIM $\uparrow$  & PSNR $\uparrow$ & SSIM $\uparrow$ 
   \\
   \hline
   \hline
PGH$^2$Net~\cite{PGH2Netisu2025prior}&\underline{41.70} &\underline{0.996}&37.52 &0.989
\\
 MAIR~\cite{MAIRjiang2026multi}
&41.48  &\underline{0.996}  &39.03 & 0.993
\\
AgenticIR~\cite{agenticir}
&41.25  &\underline{0.996}  &\underline{39.17} & 0.994
\\
HybridAgent~\cite{hybridagentli2026hybrid}
&41.52  &0.996  &39.09 &\textbf{0.996}
\\
IAMAgent~\cite{wei2026iamagent}
&41.44  &\underline{0.996}  &\underline{39.17} &\textbf{0.996} 
\\
Restore-R1~\cite{lu2026restorer1}
&41.39  &\textbf{0.997}  &39.11 &\underline{0.995} 
\\
\hline
\rowcolor{gray!20} \textbf{RecurIR(Ours)} &\textbf{41.96}	&\textbf{0.997}  &\textbf{39.19}	&\textbf{0.996}		
         \\
         \hline
    \end{tabular}}
\end{table}

\begin{table}
\centering
\caption{Image deblurring results in the task-specific setting. \label{tb:deblurgh}}
  \resizebox{\linewidth}{!}{
\begin{tabular}{ccccc}
    \hline
    \multicolumn{1}{c}{} & \multicolumn{2}{c}{GoPro}  & \multicolumn{2}{c}{HIDE} 
    \\
   Methods & PSNR $\uparrow$ & SSIM $\uparrow$ & PSNR $\uparrow$ & SSIM $\uparrow$   
    \\
    \hline\hline
    StarIR~\cite{starir11429607} & \underline{34.34}&\underline{0.970} &\underline{32.12}&\underline{0.951}
    \\
     MAIR~\cite{MAIRjiang2026multi}
&34.08  &0.965  &32.06 & 0.950
\\
AgenticIR~\cite{agenticir}
&34.18  &0.963  &\underline{32.12} & 0.949
\\
HybridAgent~\cite{hybridagentli2026hybrid}
&34.29  &0.968  &32.04 &0.950
\\
IAMAgent~\cite{wei2026iamagent}
&34.31  &\textbf{0.974}  &32.09 &\textbf{0.952} 
\\
Restore-R1~\cite{lu2026restorer1}
&34.15  &0.968  &32.11 &\underline{0.951} 
\\
\hline
\rowcolor{gray!20} \textbf{RecurIR(Ours)} &\textbf{34.81}	&\textbf{0.974}  &\textbf{32.31}	&\textbf{0.952}		
         \\
    \hline
\end{tabular}}
\end{table}

\begin{table}
    \centering
    \caption{Image denoising results in the task-specific setting.}
    \label{tab:denoisec}
    \resizebox{\linewidth}{!}{
    \begin{tabular}{c|ccccccccc}
    \hline
      & \multicolumn{3}{c}{CBSD68}  & \multicolumn{3}{c}{Kodak24} & \multicolumn{3}{c}{Urban100} 
    \\
   Methods  & 15  & 25 & 50 & 15  & 25 & 50 & 15  & 25 & 50
   \\
   \hline
   \hline
SFNet~\cite{SFNet} & 34.09 & 31.49 & 28.02 & 34.93 & 32.42 & 29.25&34.19 &32.01 & 29.03
\\
FSNet~\cite{FSNet} & 34.11 & 31.51 & 28.01 & \underline{34.95} & 32.42 & \underline{29.27}&34.15 & 32.04 & 29.15
\\
Perceive-IR~\cite{Perceive-IR10990319} &\underline{34.38} &\textbf{31.74} &28.53 &34.84 &\underline{32.50} &29.16 &34.86 &32.55 &29.42
\\
VLU-Net~\cite{VLUNetZeng_2025_CVPR}& 34.35 &\underline{31.72} &28.46 &- &- &-&\underline{34.92} &\textbf{32.71} &\textbf{29.61}
\\
     MAIR~\cite{MAIRjiang2026multi}
&34.02  &31.69 & 28.51 & 34.91 & 32.45 & 29.22 & 34.87 & \underline{32.70} & 29.49
\\
AgenticIR~\cite{agenticir}
&34.25  & 31.71 & 28.50 & 34.89 & 32.44 & 29.25 & 34.83 & 32.68 & 29.46 
\\
HybridAgent~\cite{hybridagentli2026hybrid}
&34.28  & 31.66 & 28.43 & 34.93 & \underline{32.50} & 29.23 & 34.91 & 32.68 & \underline{29.55}
\\
IAMAgent~\cite{wei2026iamagent}
&\textbf{34.40}  & \textbf{31.74} & \underline{28.65} & 34.94 & 32.41 & 29.26 & 34.90 & 32.66 & 29.49
\\
Restore-R1~\cite{lu2026restorer1}
&34.17  & 31.70 & 28.44 & 34.88 & 32.47 & 29.21 & 34.90 & 32.64 & 29.47
\\
\hline
\rowcolor{gray!20} \textbf{RecurIR(Ours)} &\textbf{34.40}	&\textbf{31.74} &\textbf{28.72} &\textbf{34.99}	&\textbf{32.52}	& \textbf{29.31} & \textbf{34.98} & 32.56 & \textbf{29.61}
         \\
    \hline
    \end{tabular}}
\end{table}

\subsubsection{Image Desnowing}

Table~\ref{tab:snow} compares the results on CSD, SRRS, and Snow100K. RecurIR achieves the highest PSNR on all three datasets. On CSD~\cite{HDCW-Netchen2021all}, it obtains 38.59 dB, slightly exceeding StarIR~\cite{starir11429607} by 0.02 dB. On SRRS~\cite{JSTASRchen2020jstasr}, RecurIR reaches 32.93 dB, improving upon the previous best result of 32.75 dB by 0.18 dB. It also achieves 34.79 dB on Snow100K~\cite{desnownet}, surpassing the strongest competing result by 0.29 dB. RecurIR additionally matches the highest reported SSIM on all three benchmarks, showing consistent performance across different snow degradation conditions.

\subsubsection{Image Dehazing}

Table~\ref{tab:sot} reports the results on SOTS-Indoor and SOTS-Outdoor~\cite{RESIDEli2018benchmarking}. RecurIR achieves the highest PSNR on both datasets, reaching 41.96 dB and 39.19 dB, respectively. On SOTS-Indoor, it outperforms PGH$^2$Net~\cite{PGH2Netisu2025prior}, the strongest competing method, by 0.26 dB. On SOTS-Outdoor, RecurIR improves upon the best competing PSNR of 39.17 dB by 0.02 dB. It also matches the highest SSIM on both datasets, indicating effective haze removal while preserving image structures.

\subsubsection{Image Deblurring}

Table~\ref{tb:deblurgh} presents quantitative comparisons on GoPro~\cite{Gopro} and HIDE~\cite{HIDE}. RecurIR achieves the best PSNR on both benchmarks, with 34.81 dB on GoPro and 32.31 dB on HIDE. Compared with StarIR~\cite{starir11429607}, the strongest competing method in terms of PSNR, RecurIR achieves improvements of 0.47 dB and 0.19 dB, respectively. Its SSIM also matches the highest reported values on both datasets, demonstrating effective blur removal and structural preservation.

\subsubsection{Image Denoising}

Table~\ref{tab:denoisec} compares denoising performance on CBSD68, Kodak24, and Urban100 under Gaussian noise levels of $\sigma=15,25,50$.
On CBSD68, RecurIR matches the highest PSNR at $\sigma=15$ and $\sigma=25$, achieving 34.40 dB and 31.74 dB, respectively. At $\sigma=50$, it obtains the best result of 28.72 dB. On Kodak24, RecurIR consistently ranks first across all three noise levels, with PSNR values of 34.99 dB, 32.52 dB, and 29.31 dB. On Urban100, it achieves the highest PSNR at $\sigma=15$ and matches the best result at $\sigma=50$, while remaining competitive at $\sigma=25$. These results show that RecurIR maintains effective denoising performance across varying noise intensities and image characteristics.

\begin{figure} % use float package if you want it here
    \centerline{\includegraphics[width=1\linewidth]{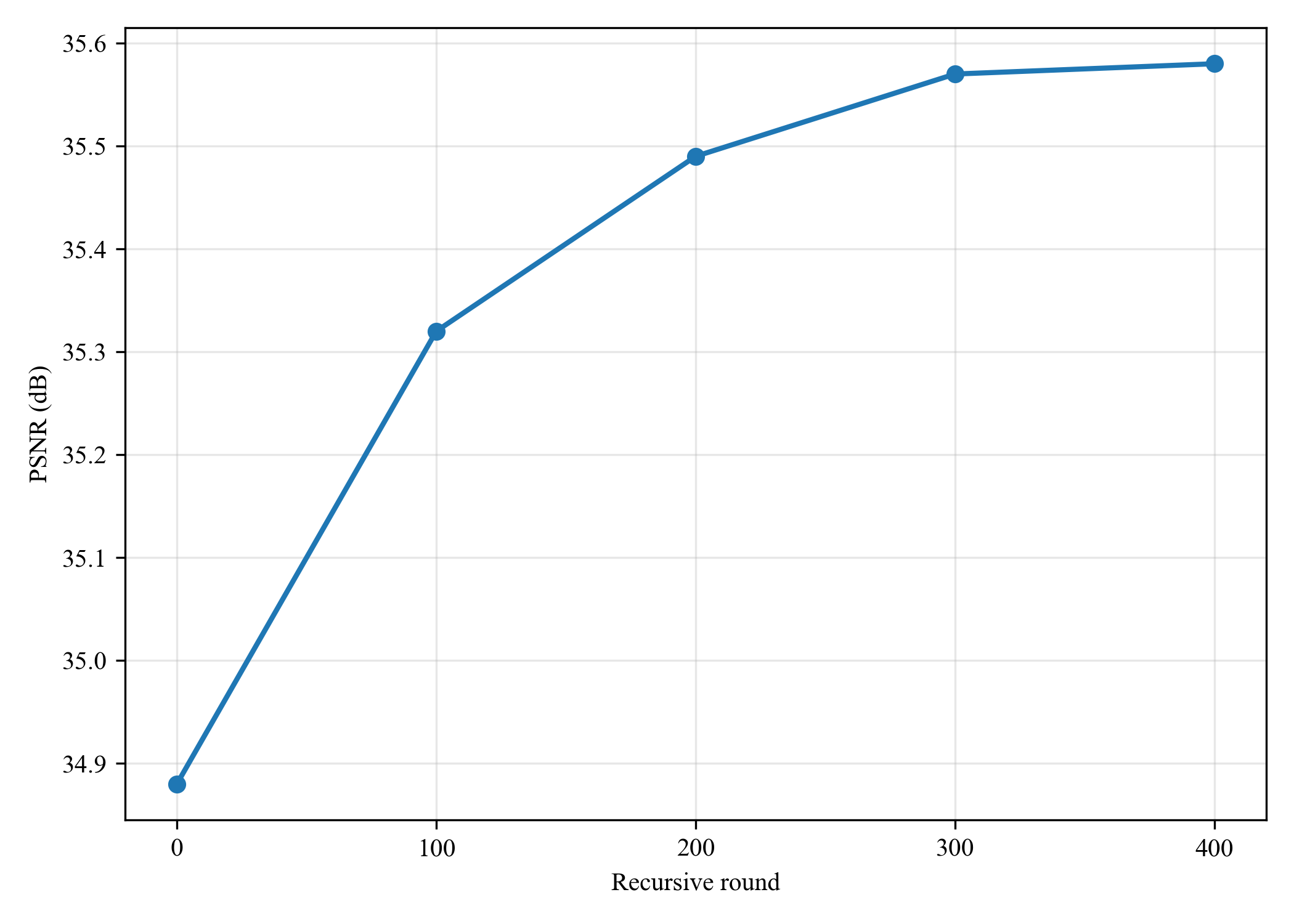}}
	\caption{PSNR performance of RecurIR over successive recursive self-improvement rounds. Restoration performance improves progressively and approaches convergence as the number of rounds increases.}
 \label{fig:recursive_rounds}
\end{figure}

\subsection{Effect of Recursive Self-Improvement}

To evaluate the effectiveness of recursive self-improvement, we examine the restoration performance of RecurIR over successive improvement rounds. As shown in Figure~\ref{fig:recursive_rounds}, the average PSNR increases from 34.88 dB at initialization to 35.32 dB after 100 rounds and further reaches 35.49 dB after 200 rounds. The improvement gradually slows, reaching 35.57 dB and 35.58 dB at 300 and 400 rounds, respectively. This trend can be attributed to the progressive refinement of decision rules during recursive learning. In the early rounds, hypothesis discovery and rule-falsification-driven task evolution identify ineffective restoration decisions and provide informative evidence for policy revision, leading to substantial performance gains. As the learned rules become more reliable and their applicability boundaries are better characterized, fewer condition changes require substantial policy adjustments, reducing the marginal benefit of subsequent revisions. Moreover, the transfer and retention verification mechanism rejects updates that offer limited improvement or compromise previously learned decisions. Consequently, performance gradually stabilizes, with an overall PSNR gain of 0.70 dB after 400 rounds.

\begin{figure}[t]
    \centering
    \includegraphics[width=\linewidth]{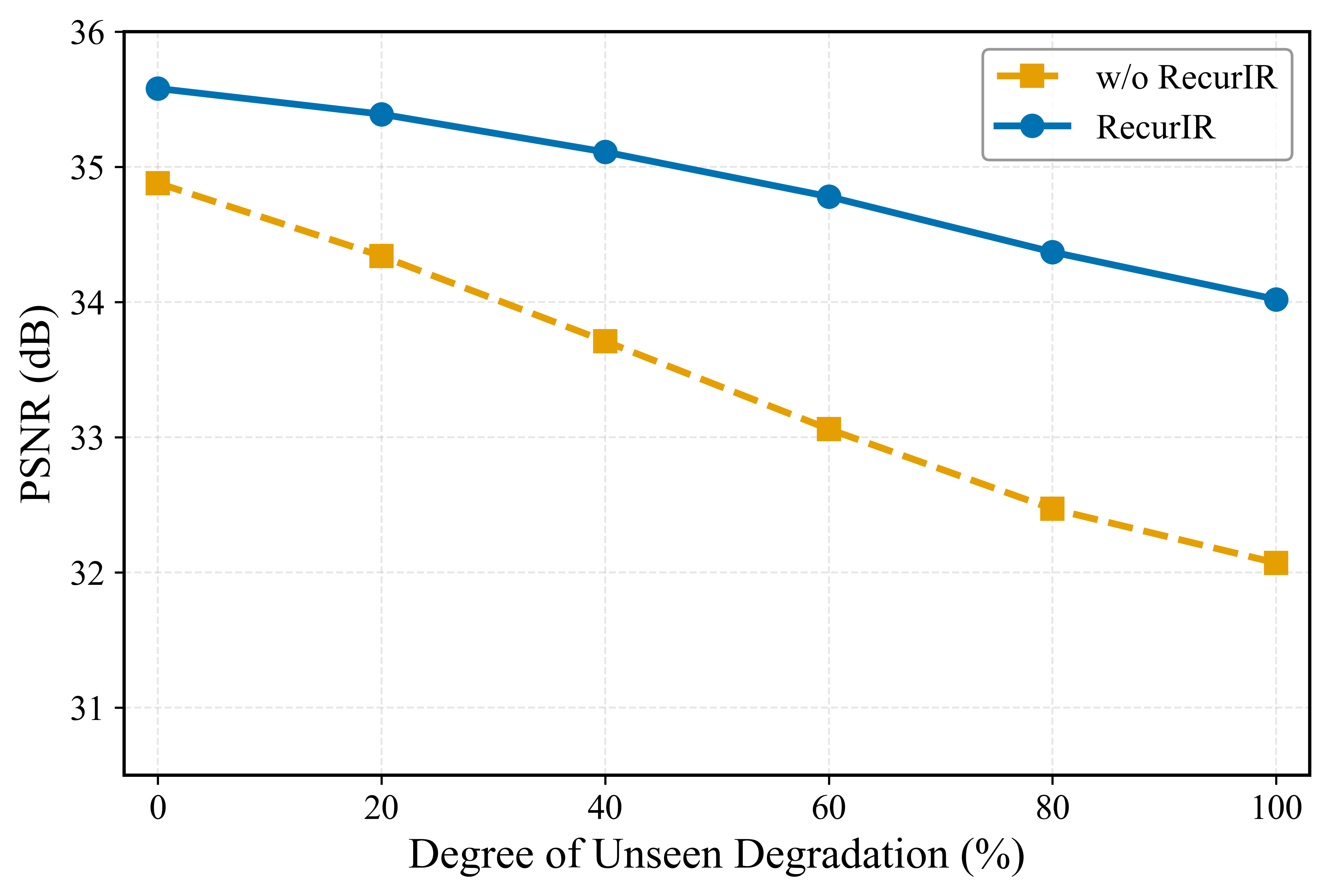}
    \caption{Restoration performance under different degrees of unseen degradation. The unseen degree reflects deviations in degradation types, severity, combinations, and spatial distributions from conditions previously encountered by the agent.}
    \label{fig:unseen_degradation}
\end{figure}

\subsection{Generalization to Unseen Degradation Conditions}

To further evaluate the adaptability of RecurIR, we examine its restoration performance under increasingly unfamiliar degradation conditions. Specifically, the degree of unseen degradation measures how much the test conditions differ from those previously encountered by the agent, including differences in degradation types, severity levels, degradation combinations, and spatial distributions. It ranges from 0\% to 100\%, with higher values indicating greater deviations from previously encountered conditions.

Figure~\ref{fig:unseen_degradation} compares RecurIR with the baseline agent without the proposed mechanisms. As the degree of unseen degradation increases, both methods exhibit a decline in PSNR. However, RecurIR consistently maintains higher restoration quality, with a smaller performance drop as the test conditions become increasingly unfamiliar. The performance gap also widens at higher novelty levels, suggesting that restoration decisions based on previously observed experiences become less reliable under substantial degradation changes. In contrast, RecurIR learns condition-dependent decision rules and revises restoration strategies according to changes in degradation characteristics. This allows the agent to adjust tool selection, execution order, operation strength, and processing regions more effectively, improving its adaptability to unseen degradation conditions.

\subsubsection{Analysis of  Restoration Results}

Table~\ref{tab:memory_comparison} illustrates the differences between conventional experience-guided restoration decisions and the proposed condition-dependent rule selection. Existing agents can reuse previously successful tool sequences based on identified degradation types. In comparison, RecurIR explicitly considers degradation severity, spatial distribution, and interactions when selecting restoration rules. As shown in the three cases, RecurIR adjusts operation strength, execution order, and processing regions according to the observed conditions, enabling more targeted restoration decisions.

\begin{table*}[t]
\centering
\caption{Illustrative comparison of restoration decisions between
existing experience-guided agents and RecurIR under different
degradation conditions. Colored text highlights degradation
severity, spatial regions, degradation interactions, and
condition-dependent restoration strategies.}
\label{tab:memory_comparison}

\footnotesize
\renewcommand{\arraystretch}{1.15}
\setlength{\tabcolsep}{6pt}

\begin{tabularx}{\textwidth}{
>{\raggedright\arraybackslash}p{0.11\textwidth}
>{\raggedright\arraybackslash}X
>{\raggedright\arraybackslash}X
}
\toprule

\textbf{Case}
&
\textbf{Existing Experience-Guided Agent}
&
\textbf{RecurIR (Ours)}
\\
\midrule

\textbf{Case 1}
\newline
Severity
Change
&
\textbf{Degradations:}
Noise and blur.

\textbf{Restoration strategy:}
Denoising $\rightarrow$ Deblurring.

\textbf{Reasoning:}
Previous restoration experience suggests removing
noise before deblurring. Therefore, the agent
selects a denoising-then-deblurring sequence.

&
\textbf{Degradation:}
The image contains \sev{moderate global blur}
and spatially varying noise.
Noise is \sev{mild in most regions} but
\sev{severe} in the
\reg{dark facade and window areas}.
\inter{Noise and blur coexist near building edges}.

\textbf{Restoration strategy:}
\rulecolor{Strong regional denoising}
in severely noisy areas,
\rulecolor{mild denoising} elsewhere,
followed by \rulecolor{global deblurring}.

\textbf{Reasoning:}
The required denoising strength varies with
local degradation severity.
The agent selects different restoration
settings to suppress noise while preserving
details in less degraded regions.

\\
\midrule

\textbf{Case 2}
\newline
Region
Change
&
\textbf{Degradations:}
Rain and haze.

\textbf{Restoration strategy:}
Dehazing $\rightarrow$ Deraining.

\textbf{Reasoning:}
Previous experience indicates that haze
removal can improve visibility before
deraining. The agent therefore follows
a dehazing-then-deraining sequence.

&
\textbf{Degradation:}
The image contains \sev{moderate global haze},
while rain streaks are concentrated in the
\reg{left and upper regions}.
The \reg{distant background} is mainly
affected by haze, whereas the
\reg{left foreground} contains
\inter{coupled rain and haze}.

\textbf{Restoration strategy:}
\rulecolor{Global dehazing}
followed by
\rulecolor{regional deraining}
restricted to rain-affected areas.

\textbf{Reasoning:}
Haze removal is required across the scene,
whereas deraining is only needed in
specific regions.
The agent adjusts the processing scope
to avoid unnecessary restoration
in unaffected areas.

\\
\midrule

\textbf{Case 3}
\newline
Mixed
Degradation
Change
&
\textbf{Degradations:}
Haze, noise, and rain.

\textbf{Restoration strategy:}
Denoising $\rightarrow$ Dehazing
$\rightarrow$ Deraining.

\textbf{Reasoning:}
Based on previous experience,
the agent removes noise first,
improves visibility through dehazing,
and then removes rain streaks.

&
\textbf{Degradation:}
The image contains \sev{mild global noise},
\sev{moderate haze}, and
\reg{localized rain in the left road area}.
The \reg{distant background} is dominated
by haze, while the \reg{dark corner regions}
contain \inter{coupled noise and haze}.
The \reg{left road area} contains
\inter{coupled rain and haze}.

\textbf{Restoration strategy:}
\rulecolor{Mild global denoising},
\rulecolor{stronger dehazing in distant regions},
and \rulecolor{regional deraining}
in rain-affected areas.

\textbf{Reasoning:}
Different regions contain different
degradation combinations and severities.
The agent selects restoration operations
and processing regions according to
their local degradation conditions.

\\
\bottomrule

\end{tabularx}
\end{table*}

\end{document}